\documentclass[journal]{IEEEtran}

\usepackage[hidelinks]{hyperref}
\usepackage{siunitx}
\usepackage{booktabs} 
\usepackage{caption} 
\usepackage{multirow}
\usepackage[caption=false]{subfig}
\usepackage{graphicx}
\usepackage{afterpage}
\usepackage{flushend}
\usepackage{amsmath}

\begin{document}

\title{On the safety of vulnerable road users by cyclist orientation detection using Deep Learning}

\author{{M. García-Venegas$^{1}$}, 
{D. A. Mercado-Ravell$^{1,2}$ } and 
{C. A. Carballo-Monsivais$^1$}
\thanks{$^{1}${Authors are with the Center for Research in Mathematics CIMAT AC, campus Zacatecas.}}
\thanks{$^{2}${D. A. Mercado-Ravell is also with Cátedras CONACYT, at CIMAT-Zacatecas}
email: {diego.mercado@cimat.mx}}}

\maketitle 

\begin{abstract}
In this work, orientation detection using Deep Learning is acknowledged for a particularly vulnerable class of road users, the cyclists.  Knowing the cyclists' orientation is of great relevance since it provides a good notion about their future trajectory, which is crucial to avoid accidents in the context of intelligent transportation systems. Using Transfer Learning with pre-trained models and TensorFlow, we present a performance comparison between the main algorithms reported in the literature for object detection, such as SSD, Faster R-CNN and R-FCN along with MobilenetV2, InceptionV2, ResNet50, ResNet101 feature extractors. Moreover, we propose multi-class detection with eight different classes according to orientations. To do so, we introduce a new dataset called ``Detect-Bike'', containing $20,229$ cyclist instances over $11,103$ images, which has been labeled based on cyclist's orientation. Then, the same Deep Learning methods used for detection are trained to determine the target's heading. Our experimental results and vast evaluation showed satisfactory performance of all of the studied methods for the cyclists and their orientation detection, especially using Faster R-CNN with ResNet50 proved to be precise but significantly slower. Meanwhile, SSD using InceptionV2 provided good trade-off between precision and execution time, and is to be preferred for real-time embedded applications.
\end{abstract}

\section{Introduction}
\label{sec:Introduction}

In recent years, significant progress has been achieved in the care and protection of Vulnerable Road Users (VRUs), including pedestrians, motorcyclists and cyclists. This effort began with the creation of road rules that attempt to convert roads into safer spaces for these users \cite{WorldHealthOrganization2018}. On the other hand, several research have been done working on vision-based detection systems for VRUs, specially pedestrians, along with improvements in the field of Intelligent Transportation Systems (ITS) for traffic monitoring and the development of Advanced Driver Assistance Systems (ADAS), such as pre-collision systems. 
However, the number of accidents and deaths on the road continues to climb at high rate, as evidenced by the World Health Organization (WHO), which ranked road traffic injuries as the tenth world's cause of death in 2002, and projected it to be the eighth cause of death by 2030 \cite{WHOProjections}. In the same way, according to WHO, there has been $1.35$ million annual deaths, it is nearly 3,700 people are dying on the world's roads every day, from where more than half of all road deaths were VRUs \cite{WorldHealthOrganization2018}, with $39\%$ pedestrians and $8\%$ cyclists. 

Accordingly, research on the detection and monitoring of pedestrians have received most of the attention \cite{HOGhuman2005,PedestrianDetec2012,HOGhuma2013,ObjectDetect2009,PedestrianDetect2013,ObjectDetec2014,lan2018pedestrian,chen2019thermal,heo2019estimation}. Unfortunately, little attention has been paid to cyclists, even though the lack of special infrastructure, protection and road safety culture makes them particularly vulnerable to road accidents. Furthermore, in contrast to pedestrian's detection, the cyclist's detection task presents other challenges, mainly due to the cyclists' visual complexity, variety of possible orientations; aspect ratios and appearance, along with the lack of labeled datasets \cite{kang2019test} and the presence of occlusions and cluttered backgrounds \cite{UnifiedCyclist2017,BenchmarkCylistDetect2016}.

Former techniques for detection of VRUs included classic artificial vision approaches for pedestrian detection, which were mainly implemented using Histogram of Gradients Oriented (HOG) for feature extraction and Support Vector Machine (SVM) for classification \cite{HOGhuman2005,PedestrianDetec2012,HOGhuma2013}. Also the Deformable Parts approach has been implemented in \cite{ObjectDetect2009} based on HOG, as part of traditional algorithms. 

\begin{figure}[!t]
\centering\includegraphics[width=0.4\textwidth]{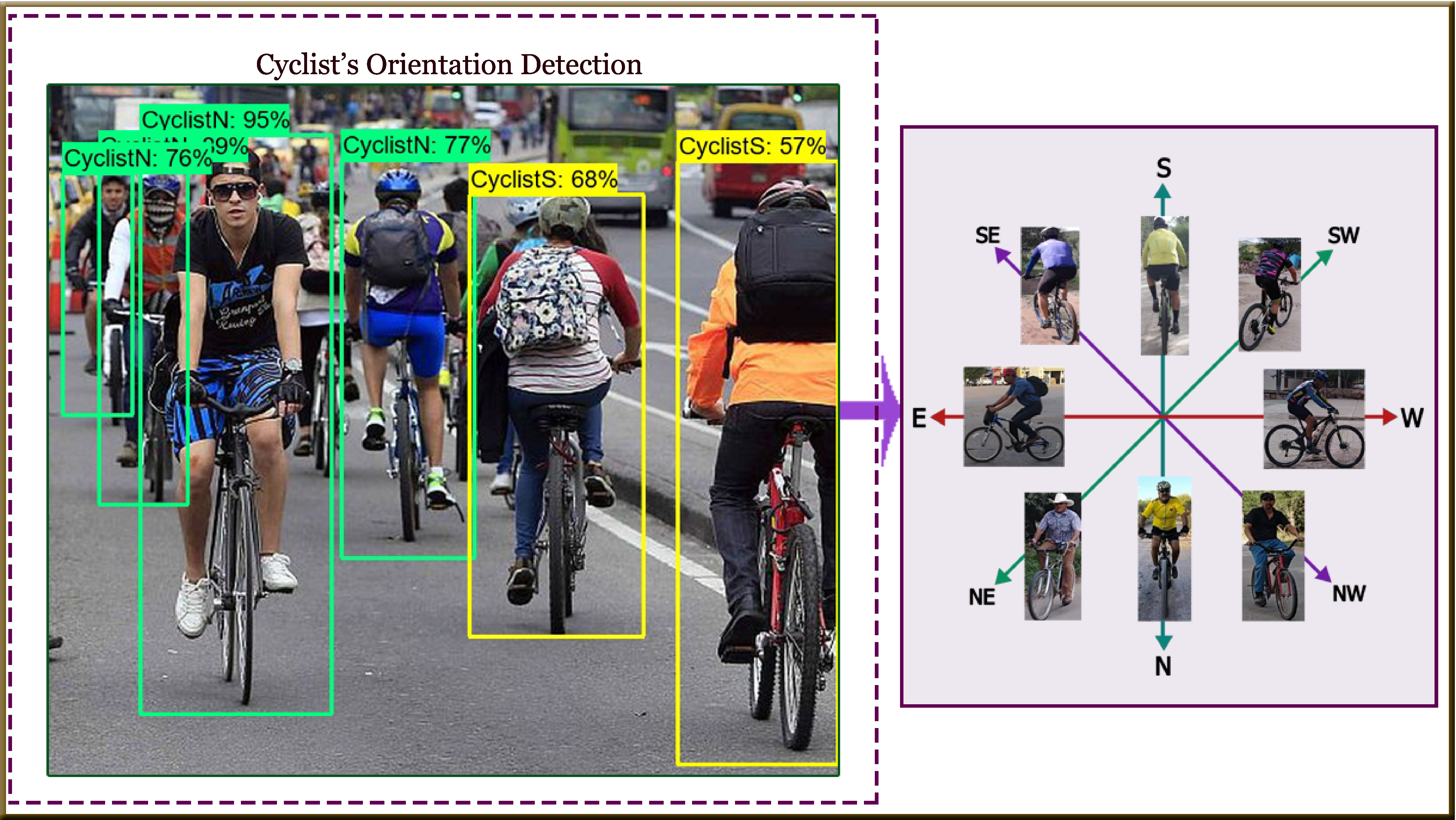}
\caption{Multi-class cyclist's orientation detection using Deep Learning. Cyclists are a particularly vulnerable kind of road user whose dynamics are tightly related to their heading. Knowing the orientation provides useful information about the future motion of the cyclist, helping to prevent accidents.}
\label{fig:0verview}
\end{figure}

On the other hand, works on bicycle detection have also been developed. For instance, in \cite{BycleDetect2015} the authors explored a combination of HOG, Shearlet Coefficient Histogram (HSC) and multi-scale local binary pattern (MLBP), using linear SVM. Moreover, in \cite{CyclistDetCrossing2010} authors used HOG features extraction with Light sampling and Pyramid sampling (HOG-LP) along with a linear SVM classifier to detect cyclists crossing a road, while in \cite{BycleHOG2012} the authors proposed the Multiple-Size Cell HOG (MSC-HOG) features detection and the RealAdaboost algorithm to detect a human on a bicycle. 

HOG-SVM combination proved to be a good option for human and bicycle detection, until the arrival of Deep Learning (DL) based algorithms within the last decade. With the boom of Convolutional Neuronal Networks (CNNs), more recent works like \cite{PedestrianDetect2013,ObjectDetec2014,lan2018pedestrian,chen2019thermal} applied them to a region proposal, with significant improvements in precision. This kind of algorithms are referred as Region-based CNN, or R-CNN.

With the rapid development in DL, more powerful machine learning methodologies and techniques have emerged, addressing the problems existing in traditional architectures. Now, with the CNNs, it is possible to learn semantic, high-level, deeper features. Taking advantage of the large learning capacity of CNNs \cite{zhao2019object}, some computer vision challenges can be considered and resolved from different viewpoints, for example, using an hierarchical feature representation. In addition, modern CNN based techniques are capable to optimize several interrelated tasks, for example to classify multiple objects at a time, or to propose multiple regions to analyze \cite{zhao2019object}. This converts them in a powerful tool for detection and classification, specially when further combined with the great advances in hardware for parallel processing, with the recent developments in Graphics Processing Units (GPUs).

More recently, for generic object detection, two types of frameworks have been introduced in the literature, ``region proposal based'' and ``regression/classification based'' \cite{zhao2019object}. The former pursues the traditional object detection pipeline, generating region proposals at first and then classifying each proposal into different object categories. These methods generally include R-CNNs \cite{ObjectDetec2014}, Spatial Pyramid Pooling Net (SSP-net), Fast R-CNN \cite{girshick2015fast}, Faster R-CNN \cite{fasterObject2015}, Region-based Fully Convolutional Network (R-FCN) \cite{dai2016r}, Feature Pyramid Networks (FPN) \cite{lin2017feature} and Mask R-CNN \cite{he2017mask}. The second framework considers object detection as a regression or classification problem, implementing a unified framework to accomplish the final result, which involves categories and locations. These methods mainly include Multibox \cite{erhan2014scalable}, AttentionNet \cite{yoo2015attentionnet}, You Only Look Once (YOLO) \cite{redmon2016you}, Single Shot MultiBox Detector (SSD) \cite{liu2016ssd}, YOLOv2 \cite{redmon2017yolo9000}, Deconvolutional Single Shot Detector (DSSD) \cite{fu2017dssd} and Deeply Supervised Object Detectors (DSOD) \cite{shen2017dsod}. In general, ``region proposal based'' methods are know to be more accurate, while ``regression/classification based'' algorithms are significantly faster \cite{zhao2019object}. 

Since the proposal of CNN have suggested a lot of improved models, including Fast R-CNN that jointly optimizes classification and bounding box regression; Faster R-CNN which introduces an additional Region Proposal Network (RPN), that can predict bounding box and score at each position simultaneously; and SSD that accomplishes object detection via regression. All of them have implemented important improvements in accuracy and execution time, and can even be used for real-time applications. In this scenario, SSD presents an interesting solution, providing the fastest detection at the cost of some precision. Then, at current state-of-the-art, there is a trade-off between precision and time response, and the best detector is to be chosen according to the application.

In consequence, CNN methods have been used for cyclist detection, being Fast R-CNN \cite{UnifiedCyclist2017,PedestrianCyclist2019}, Faster R-CNN \cite{CyclistLIDAR2017,chen2019thermal} and YOLO \cite{saranya2020cyclist,liu2019acf} the most studied ones. For example, in \cite{PedestrianCyclist2019} a unified joint detection framework for pedestrians and cyclists was presented based on Fast R-CNN to estimate three categories: pedestrians, cyclists and background, using the target candidate region selection method (MIOP) along with VGG8, VGG11 and VGG16 feature extractors. It was trained with the ``Tsinghua–Daimler Cyclist Benchmark dataset'' (TDCB) presented in \cite{BenchmarkCylistDetect2016}. This dataset has also been used in \cite{liu2019acf} where the authors proposed Aggregated Channel Feature- Region Proposal- YOLO (ACF-PR-YOLO) for cyclist detection. Meanwhile, in \cite{CyclistLIDAR2017} Faster R-CNN was used for detecting instances of cyclists in depth images, but it requires data from an extra sensor such as the laser scanner. It was showed that this method outperforms the classical HOG+SVM object localization on the synthetic depth images dataset. In addition, in \cite{chen2019thermal} the authors evaluated the pedestrian and cyclist detection using thermal images. Also, in most recent investigations, \cite{saranya2020cyclist} studied cyclist detection using the Tiny YOLO v2 algorithm with a dataset proposed on \cite{BenchmarkCylistDetect2016}. 

As can be seen, most of the studies have focused only in the cyclist detection task. Nevertheless, in the context of ITS and VRUs' safety, detecting the objects of interest is not enough, provided that VRUs are constantly moving and changing their appearance. It is then of great interest to further gather information about the movement of the cyclist, and try to predict their location in the near future to determine on-time weather or not it will be in danger. While pedestrians may be unpredictable in the direction of their movement, both cyclists and motorcyclists are a different matter, provided that they always move in the forward direction. Then, knowing the orientation of this kind of VRUs is of great relevance, since it provides an important notion about their future movement, which may significantly help to avoid accidents.

In this work, we are interested in the road safety of a particular kind of VRU whose dynamics strictly depend on their heading, such as two-wheeled vehicles like bicycles. Henceforth, we propose a multi-class detection strategy based on the cyclist orientation (see Fig. \ref{fig:0verview}). Accordingly, we introduced a new dataset called "Detect-Bike", containing $20,229$ cyclist instances over $11,103$ images, which has been labeled based on the cyclists orientation. In order to accomplish orientation detection, we make use of the state-of-the-art DL techniques, such as, Single Shot Multibox Detector (SSD), Faster Region-based Convolutional Network (Faster R-CNN) and Region-based Fully Convolutional Networks (R-FCN) meta-architectures in combination with MobilenetV2, InceptionV2, Residual Network 50-layers (ResNet50), Residual Network 101-layers (ResNet101) and InceptionResNetV2 feature extractors. Taking advantage of Transfer Learning with pre-trained models and TensorFlow Object Detection API (Application Programming Interface), we implemented different models for cyclist's orientation detection and evaluate them thoroughly. Experimental results suggested that Faster R-CNN with InceptionV2 offers the best alternative for the cyclists detection task when greater precision is required, however, when a better time response is needed, the best option is given by SSD with InceptionV2. Finally, for cyclist's orientation, Faster R-CNN with ResNet50 was superior in precision, but like for cyclists detection Faster R-CNN with InceptionV2 provided a better trade-off between precision and time response, while again SSD with InceptionV2 was the fastest solution with acceptable precision.

The remaining of the paper is organized as follows: Section \ref{sec:relatedWorks} presents related works about cyclist's orientation detection. Afterwards, Section \ref{sec:methodology} describes the proposed methodology introducing a new cyclist detection dataset. Then, Section \ref{sec:exp} describes the evaluation protocol and comparison results for cyclist detection and orientation detection. Finally, Section \ref{sec:Conclusion} discusses the conclusions and future work.

\section{Related works}
\label{sec:relatedWorks}

As previously stated, it is not enough to merely detect objects in the context of VRUs' safety and ITS, but it is of vital importance to further know their dynamics in order to predict their position in time, henceforth, detect potential collision danger. For the particular case of two-wheeled vehicles, such as bicycles, which always move forward in normal conditions, it is very interesting to know their orientation, since it provides great insight on their movement. Unfortunately, little attention have been payed to this key task, and to our knowledge there are not many works reported in the literature that attempt to detect the cyclist orientation. 

Somehow related, there are some studies that have shown interest in the division of the cyclist from its aspect ratio in order to facilitate the cyclist detection \cite{BenchmarkCylistDetect2016,UnifiedCyclist2017}. In \cite{BenchmarkCylistDetect2016} a new method called Stereo-Proposal based Fast R-CNN (SP-FRCN) was introduced to detect cyclists using their own dataset TDCB, which contains VRUs including pedestrians, cyclists and motorcycles instances, recorded from a moving vehicle in the urban traffic of Beijing. It divides the cyclist samples into three classes: narrow, intermediate and wide, based on the aspect ratio of bikes. They consider three difficulty levels (easy, moderate and hard) according to the object size and occlusion level. In a similar fashion, in \cite{UnifiedCyclist2017} the same authors presented another unified framework for concurrent pedestrian and cyclists detection, including a proposal method called Upper Body - Multiple Potential Regions (UB-MPR) for generating object candidates and using Fast R-CNN for classification and localization. Even though the aspect ratio is directly related to the cyclist orientation, such works do not intend to detect it, neither they are interested in predicting the cyclists movement.

Regardless of the importance of detecting both the position and orientation in order to predict potential accidents, orientation detection is rarely considered in cyclist's detection \cite{tian2017detection}. In \cite{tian2015bfast} the authors proposed dividing the cyclists into eight subcategories based on orientation using the KITTI dataset \cite{geiger2012we}, analogously to an idea previously used for vehicle detection \cite{subcategories}. For each orientation a detector is built in a cascaded structure, using a classical approach with HOG features, along with Decision Trees (DT) and one SVM for both visible and occluded cyclists. Besides, they used a geometric method for Region of Interest (RoI) extraction and Kalman Filters to estimate cyclists’ trajectories, with a total of 16 detectors, considerably increasing the computational complexity of the algorithms. In a posterior work in \cite{tian2015afast}, the same authors proposed using Decision Forest (DF) instead of DT, which along with the inclusion of a non-maximum suppression algorithm allowed them to reduce to half the number of cascaded detectors. Lastly, in \cite{tian2017detection} the same authors improved their previous proposal by adding a max pooling operation over spatial bins and orientation channels. Nevertheless, the use of multiple detectors and traditional techniques significantly compromises precision and increases the amount of calculations required, when compared to modern CNN based techniques.

In the growing interest of protecting VRUs and not only knowing their location, but also their orientation, the KITTI dataset continues to be a benchmark for current work \cite{chen20153d,guindel2017modeling,guindel2019traffic}. This dataset provides 3D bounding box annotations, for object classes such as cars, vans, trucks, pedestrians, cyclists and trams, and it is evaluated in three regimes: easy, moderate and hard, depending on the levels of occlusion and truncation. For example, in \cite{guindel2018fast} the authors proposed a joint detection and viewpoint estimation system with a monocular camera using Faster R-CNN meta-architecture with VGG16 feature extractor, for determining the orientation of the three objects: car, pedestrian and cyclist. For estimation of the object's viewpoint they adopt discrete pose estimators to partition the view sphere into a predefined number of bins, and compute the viewpoint as the weighted average of adjacent viewpoint bin centers, using their respective estimated probabilities provided by the network. In a posterior work in \cite{guindel2019traffic}, the same authors proposed an approach for recognition and 3D localization of dynamic objects on images from a stereo camera, with a stereo-based 3D reconstruction of the environment, besides they evaluated with others feature extractors, such as, Zeiler and Fergus (ZF) and MobileNet using the KITTI dataset with $1,626$ samples of cyclist, finally they implemented their system on an intelligent vehicle. Unfortunately, it is not possible to make a direct comparison with these works, because the approach adopted is different, they perform the detection and estimation of the viewpoint using inference algorithms, while we divided each orientation as an independent class, so we avoid using estimation algorithms, then the task of training the neural network with eight predefined classes allows us to directly solve the problem as a multi-class detection task in monocular images.

As can be seen, nowadays it is not only sufficient to carry out object detection, but it is also necessary to include their orientation. In this case, the trend of the most recent works is the use of 3D stereo vision and continue to take advantage of the KITTI dataset. Lamentably, this database only provides a small amount of cyclist instances (no more than $2,000$ \cite{BenchmarkCylistDetect2016,guindel2019traffic}). From our part, we only focus on monocular vision with a new dataset specialized in cyclists, that has been labeled with the orientation and has been tested thoroughly, comparing the main techniques reported in the literature that have proven to be more efficient for this problem, hence offering an update on the evaluation of the state-of-the art.

Henceforth, we present a new cyclist dataset called ``Detect-Bike'', annotated according to the cyclist orientation, which further takes into consideration particular aspects from our local context. We believe that current available databases do not take into account particular characteristics which are common to certain regions in the world, which is the case of the Mexican countryside, where people normally use traditional clothing and hats, which may compromise the performance of the algorithms. This provides a benchmark for cyclist's orientation detection, which consists of two subsets: ``Detect-Bikev1'' with bounding box based labels that provides the class: Cyclist, and ``Detect-Bikev2'' with bounding box based labels according to eight different classes depending on the orientation. These subsets stress the importance of taking into consideration the cyclist direction of movement in the context of intelligent vehicles. Besides, we present an in depth evaluation of the most important meta-architectures up to date, such as SSD, Faster R-CNN and R-FCN, along with the most relevant feature extractors like MobilenetV2, InceptionV2, ResNet50, ResNet101. Finally, we discuss the main results observed in the evaluation and propose our choice of the best models under various scenarios, according to different criteria such as precision and time response, as well as a good trade-off between them.

The main contributions of this paper are then summarized as follows:
\begin{itemize}
  \item Creation of a new database which has been labeled based on the cyclist's orientation, which contains $20,229$ cyclist instances over $11,103$ images. 
  \item An alternative technique for cyclist's orientation detection in monocular images, which allows us to have a good notion of the cyclist's movement,  which is of great relevance to prevent accidents.
  \item A more in depth and updated evaluation of state-of-the-art techniques to perform this cyclist detection task. Particularly to determine a good trade-off between precision and time response under different scenarios.
\end{itemize}

\section{Methodology}
\label{sec:methodology}
\begin{figure}[!t]
\centering\includegraphics[width=2.5in]{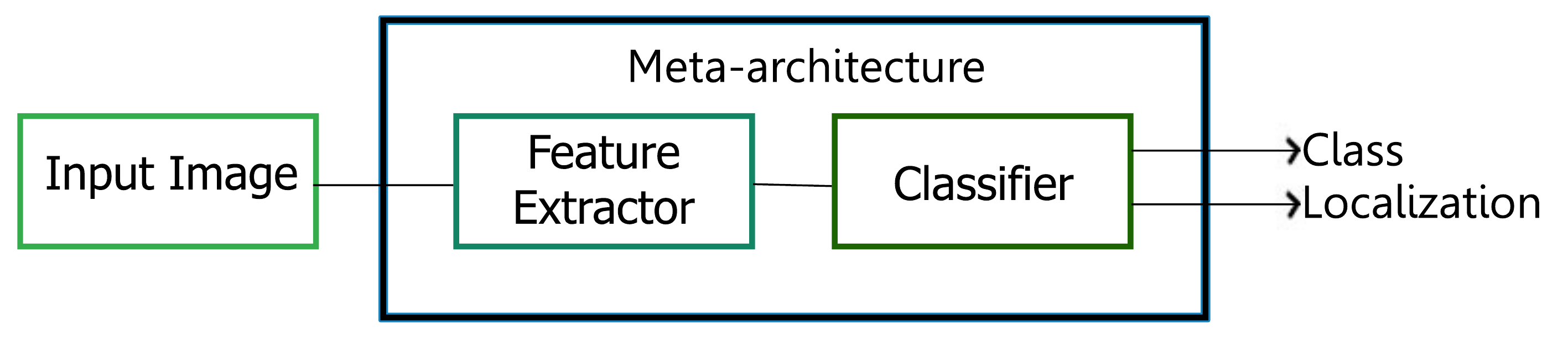}
\caption{Object detection model. A meta-architecture is composed of any convolutional feature extractor to obtain high level features \cite{huang2017speed}, then, this features are used for classification, providing the class and the localization of the bounding box of the cyclists in the image.}
\label{fig:ObjectDetectionModel}
\end{figure}

We propose cyclist detection and orientation estimation using the meta-architectures available on \cite{huang2017speed}, which consist of a single convolutional network, trained with a mixed regression and classification objective, and use sliding window style predictions, as shown in Fig. \ref{fig:ObjectDetectionModel}. In this work,  SSD \cite{liu2016ssd}, Faster R-CNN \cite{fasterObject2015}, R-FCN \cite{dai2016r} meta-architectures have been evaluated in combination with the state-of-the-art feature extractors MobilenetV2 \cite{sandler2018mobilenetv2}, InceptionV2 \cite{szegedy2016re},  ResNet50, ResNet101 \cite{he2016deep} and InceptionResNetv2 \cite{szegedy2017inception}. Moreover, we introduce a new dataset for training and testing called ``Detect-Bike'', containing images with cyclist instances taken from our nearby environment in the central-north region in Mexico, considering the particular aspects of our region such as people with traditional hats. The new dataset has been labeled according to the cyclist orientation, into eight different categories. We consider two different cases for object detection using deep learning. First, single class detection for the class cyclist. Second, multi-class detection with eight classes of cyclist's orientation.  As part of the methodology we performed the implementation in the Tensorflow Object Detection API, taking advantage of the pre-trained models provided on \cite{TensorFlow2019} for Transfer Learning.

\begin{figure}[t!]
\begin{center}
  \subfloat[][Training pipeline]{
   \label{f:Solution}
    \includegraphics[width=0.35\textwidth]{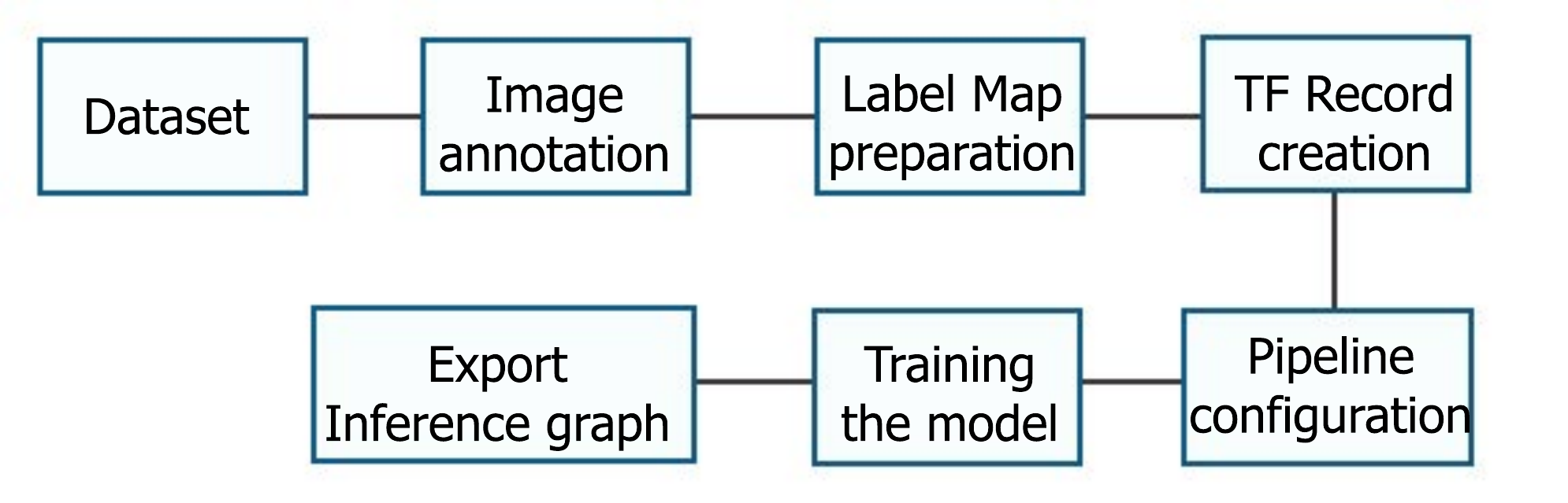}}
    \newline
  \subfloat[][Inference pipeline]{
   \label{f:Visual}
    \includegraphics[width=0.26\textwidth]{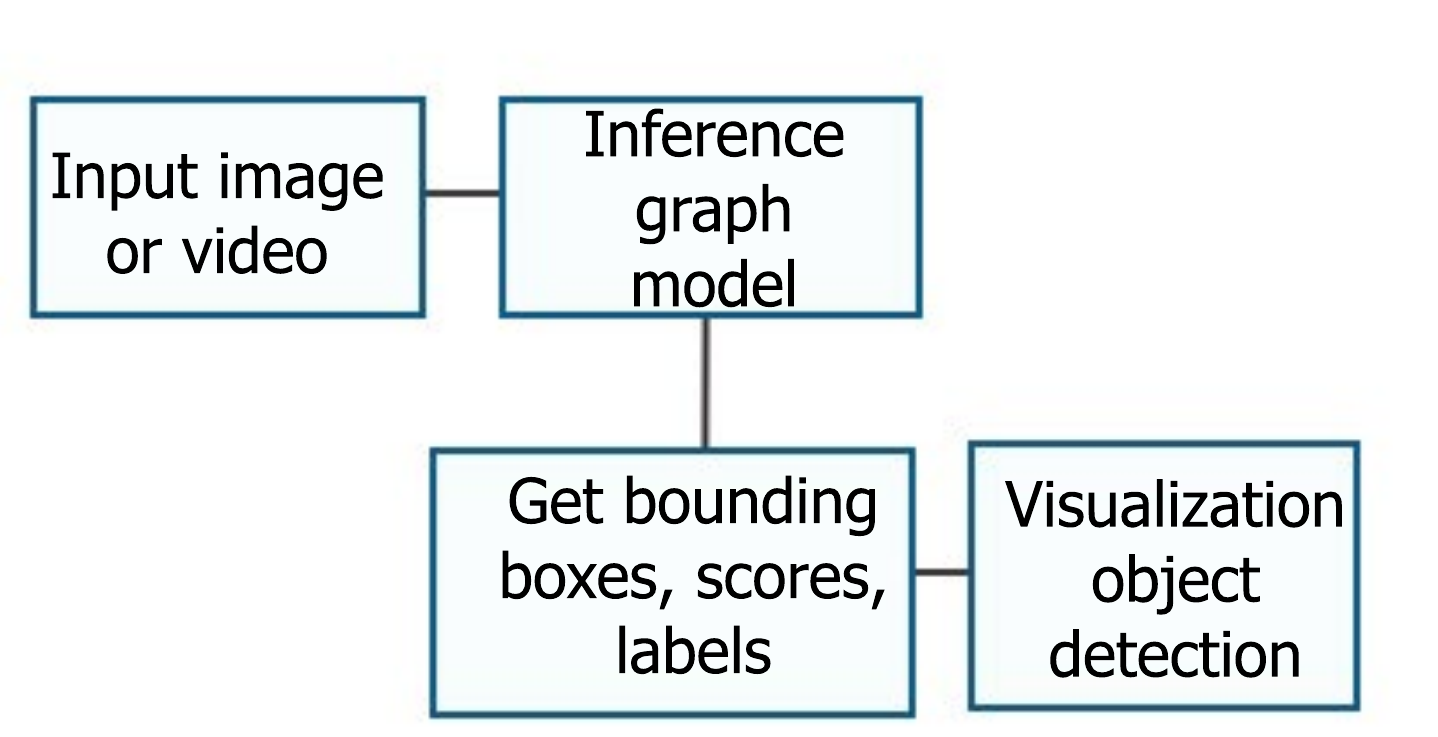}}
 \caption{Workflow diagram. a) As input for training, every image has been labeled to obtain the ground truth of each of the instances of cyclists, also TF Record files and the label map has been configured into the pipeline, and as output it obtains the inference model. b) An input image or video is evaluated using the inference model, which provides the bounding boxes, scores and classes.}
 \label{fig:SystemOverview}
\end{center}
\end{figure}

The workflow diagram shown in Fig. \ref{fig:SystemOverview} is divided into two phases, the first is the model's training phase and the second  is the model's inference where the visual results are shown. Fig. \ref{f:Solution} depicts the workflow for the model training phase, where first of all the new cyclists' dataset was generated and labeled with the orientations. Within the Label Map file, each of the classes has been indicated, and for the generation of the binary files TF Records, each bounding box of each instance is taken for each class. For the pipeline configuration, the parameters for executing the detector have been established. Here the paths for checkpoints, label map and TF Record have been settled, and finally the frozen model has been generated. In Fig. \ref{f:Visual} when the training model has been frozen, it is possible to execute the model for detection, in this case the input can be an image or video. Then, the model inference graph displays the bounding boxes, scores and labels of each class. In order to display the detected cyclists.

\subsection{Cyclist Image Dataset}
\begin{figure}[t!]
\begin{center}
  \includegraphics[width=0.35\textwidth]{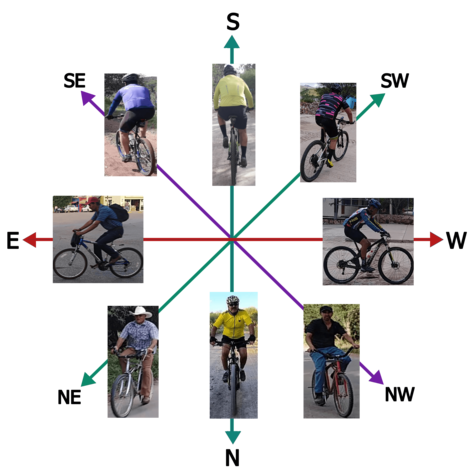}
\caption{Cyclists are divided into 8 classes according to orientation: CyclistN, CyclistNE, cyclistE, cyclistSE, cyclistS, cyclistSW, cyclistW and cyclistNW. }
\label{fig:DiagramCyclist}       
\end{center}
\end{figure}

\begin{table}[hb!]
\begin{minipage}[c]{0.45\textwidth}
\caption{Detect-Bikev2 database for cyclists' orientation. At current time we provide $20,229$ instances over $11,103$ images, where $80\%$ of the images were used for the training set and $20\%$ for the test set. By now, we focus mainly in large size instances.
\label{table:CyclistDBOrientation}}
\end{minipage}
\addtolength{\tabcolsep}{-1pt}
\begin{center}
\begin{tabular}{llll}
\toprule
{Class Orientation}& {Total} & {Training} & {Test}  \\
\midrule
{Cyclist\textbf{N}} & 3,870 & 3,100 &   770 \\\hline
{Cyclist\textbf{NE}} & 3,023 & 2,406 &  617 \\\hline
{Cyclist\textbf{E}} & 2,232  &  1,789 &  443  \\\hline
{Cyclist\textbf{SE}} & 1,849  &   1,513 &  336  \\\hline
{Cyclist\textbf{S}} & 2,427  &  1,931 & 496    \\ \hline
{Cyclist\textbf{SW}} &  1,864 &  1,478 &   386  \\\hline
{Cyclist\textbf{W}} & 1,918  & 1,544 &   374 \\\hline
{Cyclist\textbf{NW} }& 3,046 & 2,438 &  608  \\\hline
{\textbf{Total}} &\textbf{20,229}  & \textbf{16,199} & \textbf{4,030} \\\hline
\bottomrule
\end{tabular}
\end{center}
\end{table}

The database is essential for the proper training of object detectors. As we have seen, the first step in building an object detector is the preparation of a dataset with labeled images. In the case of public databases available with cyclist's instances, only two databases were found: TDCB \cite{BenchmarkCylistDetect2016}, which has been considered for the detection of pedestrians and cyclists mainly, however it is not labeled based on the orientation of the cyclist, and KITTI database \cite{geiger2012we}, which unfortunately provides a very limited number of cyclist's instances (less than $2,000$). For this reason, we propose a new dataset with $20,229$ instances over $11,103$ images, labeled according to eight different classes of  orientation, as observed in Fig. \ref{fig:DiagramCyclist}. Moreover, we included images taken from our surroundings, since the available datasets do not represent properly the particular problematic present in our national and regional context, as is the case for several other regions in the world, such as people using traditional clothing like hats. Furthermore, we provide new labels according to the cyclist orientation, which may be used to determine their heading and predict their movement, which is vital to prevent accidents on the roads. 

For the evaluation of meta-architectures, we have divided our own dataset ``Detect-Bike'' into two subsets: Detect-Bikev1 and Detect-Bikev2, whereas the former contains $12,075$ cyclist instances over $6,605$ images and is labeled for detection of a single class, Cyclists. The images were collected in approximately $450$ videos and images taken from sports events and streets in the state of Zacatecas Mexico, where some people usually wear traditional hats to protect themselves from the harsh sun. On the other hand, Detect-Bikev2 combines these images with others obtained from the web, and labeled according to our approach for multi-class detection depending on the cyclist's orientation.  For this approach, in total we annotated $20,229$ cyclist instances over $11,103$ images. In both cases, $80\%$ of the images were used for the training set and $20\%$ for the test set. For dataset Detect-Bikev2, all images were labeled and divided into eight classes. As suggested in \cite{guindel2018fast}, eight categories are a good choice to represent the cyclist's orientation, similarly to \cite{tian2015bfast}, with the difference that a special label is assigned according to the compass rose, it is, according to the orientation of the cyclist, in order to know the direction of movement of the cyclist. These classes are then Cyclist N, Cyclist NE, Cyclist E, Cyclist E, Cyclist SE, Cyclist S, Cyclist SW, Cyclist W and Cyclist NW, as shown in Fig. \ref{fig:DiagramCyclist}. Meanwhile, Table \ref{table:CyclistDBOrientation} shows the number of each class instance available in the dataset.

\subsection{Convolutional Neural Networks}
 \begin{figure*}[ht!]
\begin{center}
  \includegraphics[width=0.75\textwidth]{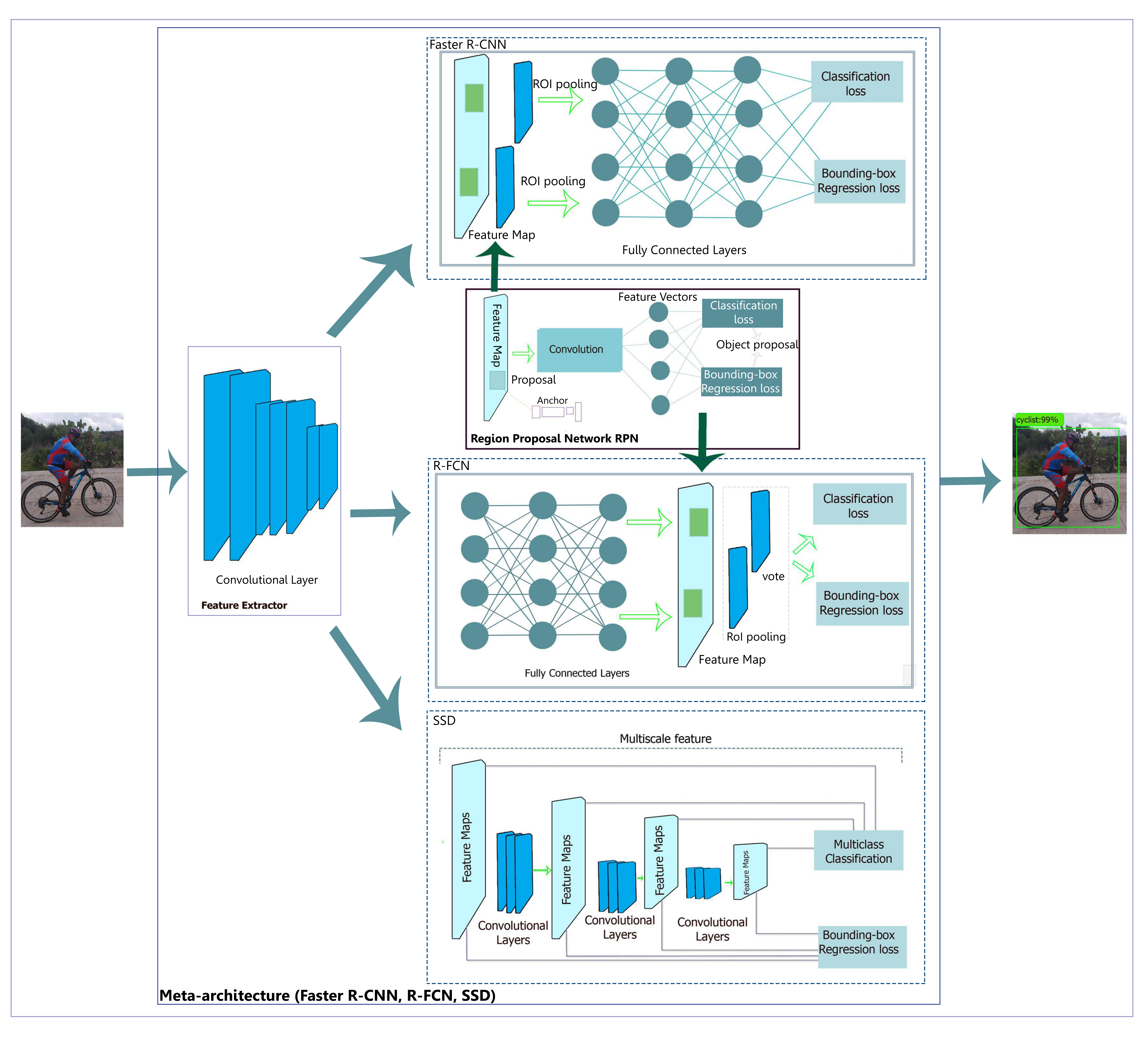}
\caption{Overall strategy. First an input image is needed, as part of the meta-architecture one of the feature extractors is chosen: MobileNetV2, InceptionV2,  ResNet50, ResNet101 or InceptionResNetV2. Then, we can notice that both Faster R-CNN and R-FCN use RPN to generate object proposals, and are well known for their superior precision. On the other hand SSD uses multi-scale feature maps for detection in a single stage, considerably reducing the execution time.}
\label{fig:DiagramArch}       
\end{center}
\end{figure*}

Deep Learning, and more in particular CNN based methods are considered to be the best for object detection up to today \cite{ObjectDetec2014}, \cite{zhao2019object}. Hence, in this work we use them for the cyclist detection task. In order to establish a baseline comparison between the main object detection algorithms for this particular task, we have studied three of the main meta-architectures reported in the literature, such as Faster-RCNN \cite{fasterObject2015}, R-FCN \cite{dai2016r}, and SSD \cite{liu2016ssd}. Since Faster-RCNN and R-FCN are region proposal based, they are well known for their superior precision, while SSD is regression/classification based, therefore it provides considerably faster response, at the cost of precision. We are particularly interested in finding out a good trade-off between precision and detection time. 

Faster R-CNN \cite{fasterObject2015} performs the detection in two stages: a Region Proposal Network (RPN) produces a set of object proposals on an input image, each object proposal with an objectness score, and a network of object detection that uses the proposals to detect the classes. Similarly, the R-FCN method employs two-stages for object detection: region proposal and region classification, but the ROIs are taken from the last layer of features before the prediction, instead of taking them from the same layer where region proposals are predicted, it is to say, the main difference being the order in which the Fully Connected Layers (FCL) are applied. Besides R-FCN applies a voting method to detect the object. Both Faster R-CNN and R-FCN use a RPN to obtain the localization loss of the bounding box regressor for the RPN, and objectness loss, it is, if a bounding box corresponds to an object of interest or is part of the background. All detectors obtain both classification and localization loss for the final classifier.

On the other hand, Single Shot Detector (SSD) \cite{liu2016ssd} works using a single deep neural network for detecting multiple objects within an image, without requiring a second stage, but combining ideas from RPN in Faster R-CNN, where SSD simultaneously produces a score for every category for each object. Such strategy allows it to considerably reduce the execution time, at the cost of precision, making it an interesting alternative for real-time applications. This meta-architecture performs detection over multiple scales, by operating on multiple convolutional feature maps, adding feature layers that decrease in size progressively, allowing prediction of detections at multiple scales, where each added feature layer can produce a fixed set of detections, using a set of convolutional filters, each of which predicts category scores and box offsets for bounding boxes of appropriate sizes. 

In all meta-architectures, first, the images are processed by a feature extractor to obtain high level features. The choice of the feature extractor is very important, since the number of parameters and types of layers directly affect memory usage, time response, complexity and performance of the detector \cite{huang2017speed}. In this paper, five state-of-art feature extractors are considered MobilenetV2, InceptionV2, ResNet50, ResNet101 and InceptionResNetV2, provided that they have been efficient for the task of object detection, and they are available with the Open Source TensorFlow Object Detection API. 

The MobileNetV2 \cite{sandler2018mobilenetv2} structure is built on depth-wise separable convolutions, where a full convolutional operator is replaced with a factorized version that splits convolution into two separate layers, the first layer called a depth-wise convolution, which applies a single convolutional filter per input channel. The second layer is a point-wise convolution with a $1\times1$ kernel, which builds new features through computing linear combinations of the input channels. This feature extractor improves the state-of-the-art performance of mobile models on multiple tasks and benchmarks.

Other interesting extractor is called Inception \cite{szegedy2015going}, which is based on finding how an optimal local sparse structure, in a convolutional vision network, can be approximated and covered by dense components, assuming that each unit from the earlier layer corresponds to some region of the input image, and grouping these units into filter banks. One of the main beneficial aspects of this architecture is that it allows for significantly increasing the number of units at each stage, without a heavy increase of computational complexity. This enables the creation of lighter versions, such as, InceptionV2 and InceptionV3 that were introduced in \cite{szegedy2016re}. In InceptionV2 convolution factorization was added and filter banks were expanded as new variant of the Inception network, which yields a good speedup in training. At current time, the 4th version InceptionV4 has been released, which was introduced in the same work that Inception ResNetV2 \cite{szegedy2017inception}.

In ResNet \cite{he2016deep}, a deep residual learning is introduced, such that it allows that the learnable parameters of a layer or set of layers in a CNN, are mapped into a residual function. This eases the optimization by providing faster convergence at the early stage. ResNet has been divided into 18, 34, 50, 101 and 152 layers, where increasing the number of layers increases the depth, however among 50, 101 and 152 layers there is no significant increase in accuracy, and even for the 152-layer ResNet has lower complexity than VGG-16. ResNet won the 1st place on the tasks of ImageNet detection, ImageNet localization, COCO detection and COCO segmentation in 2015.

Inception ResNetV2 \cite{szegedy2017inception} is a hybrid feature extractor that combines the Inception style networks and utilize residual connections instead of filter concatenation, with an important improvement in the recognition performance. In comparison with InceptionV4, Inception ResNetV2 obtains similar accuracy, nonetheless is faster because InceptionV4 contains higher number of layers.

As mentioned before, cutting edge meta-architectures for object detection require a feature extractor and a classifier, the Fig. \ref{fig:DiagramArch} summarizes the overall methodology employed in this work.

We have selected these meta-architectures and feature extractors because they have achieved state-of-the-art detection performance and are commonly employed in Generic Object Detection. In addition, the combination of these models has not been recorded in the literature for the detection of cyclists as a class, nor for the detection of their orientation.

\section{Experiments and results}
\label{sec:exp}
\begin{figure*}[ht!]
\begin{center}
\begin{tabular}{c c c}
\includegraphics[width=0.27\textwidth]{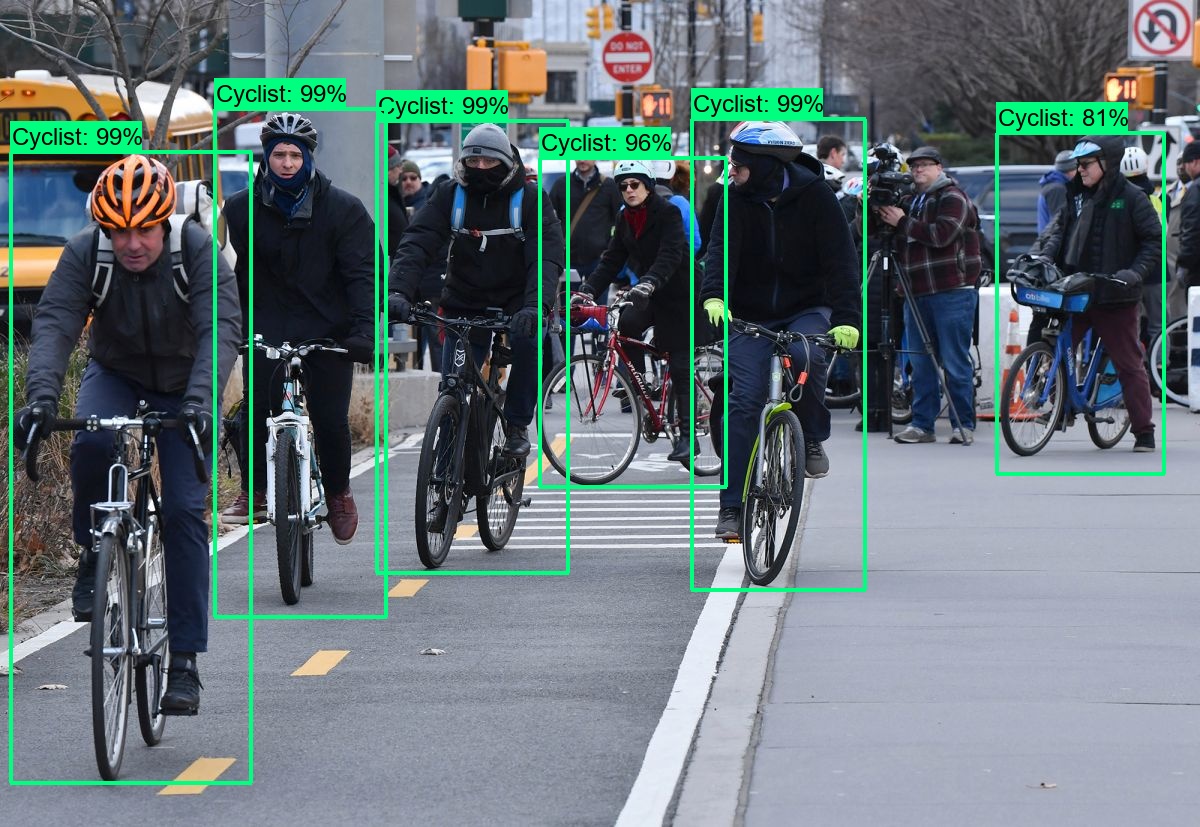} & \includegraphics[width=0.27\textwidth]{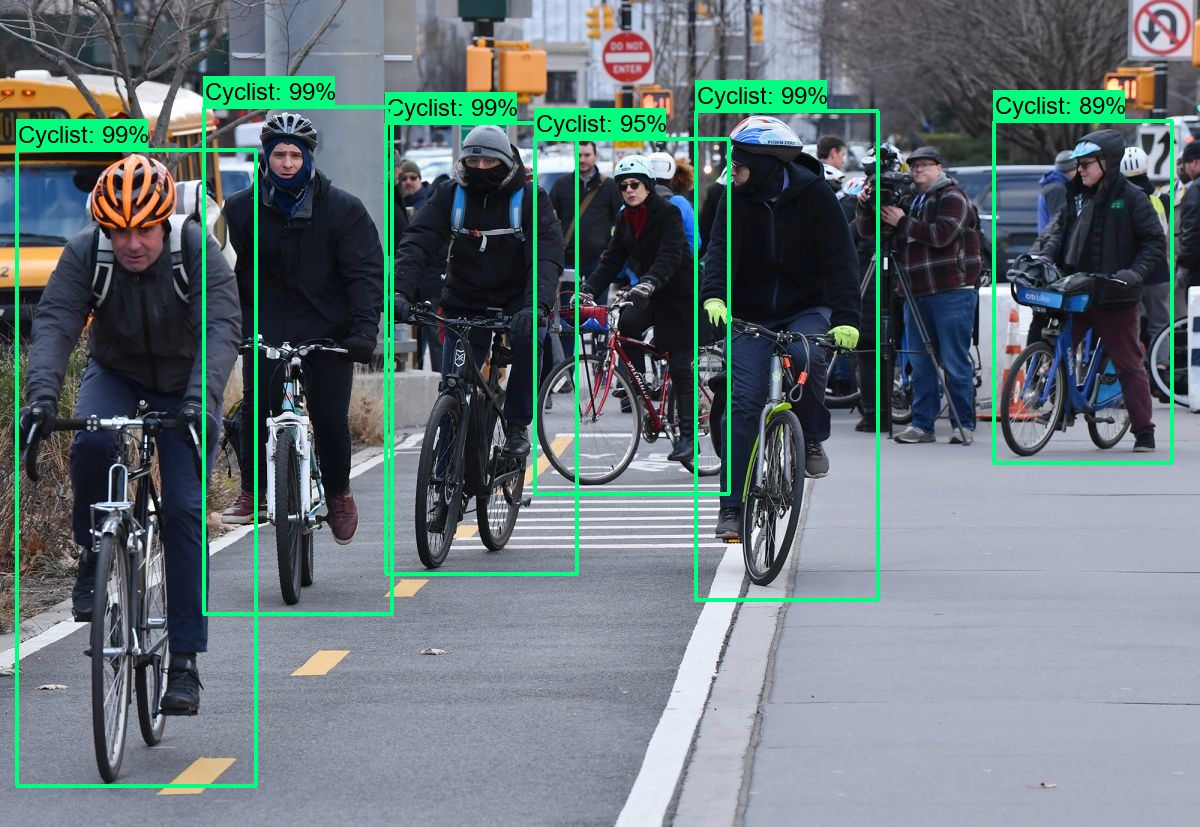} & \includegraphics[width=0.27\textwidth]{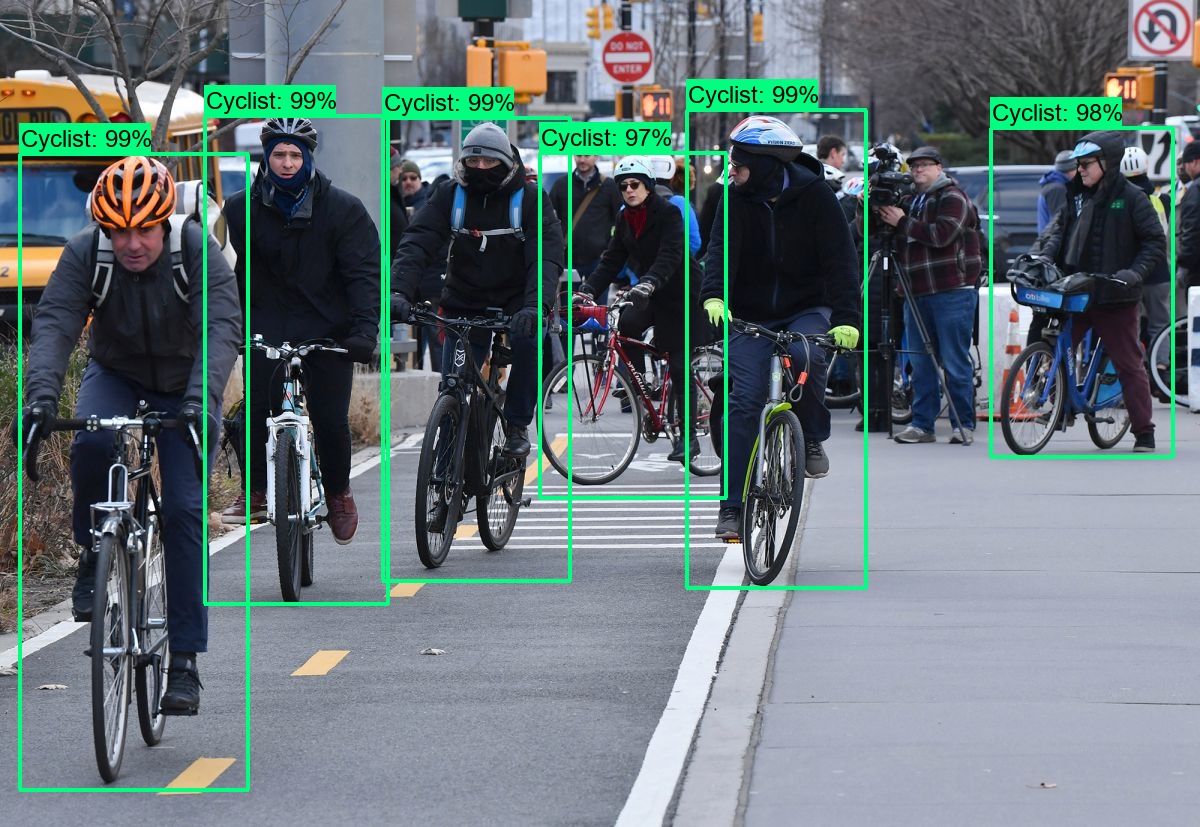} \\
(a) FasterRCNN-ResNet101 & (b) FasterRCNN-InceptionV2 & (c) FasterRCNN-ResNet50 \\[6pt]
  \includegraphics[width=0.27\textwidth]{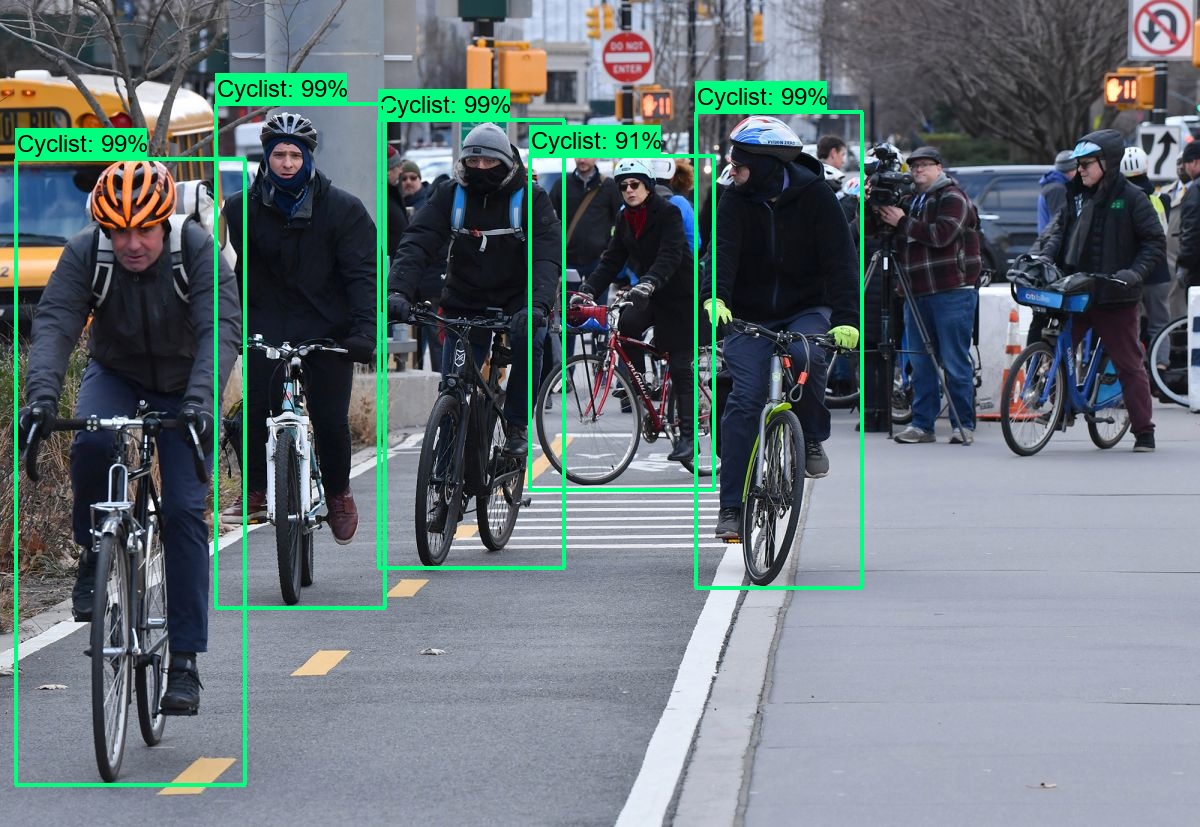} &  \includegraphics[width=0.27\textwidth]{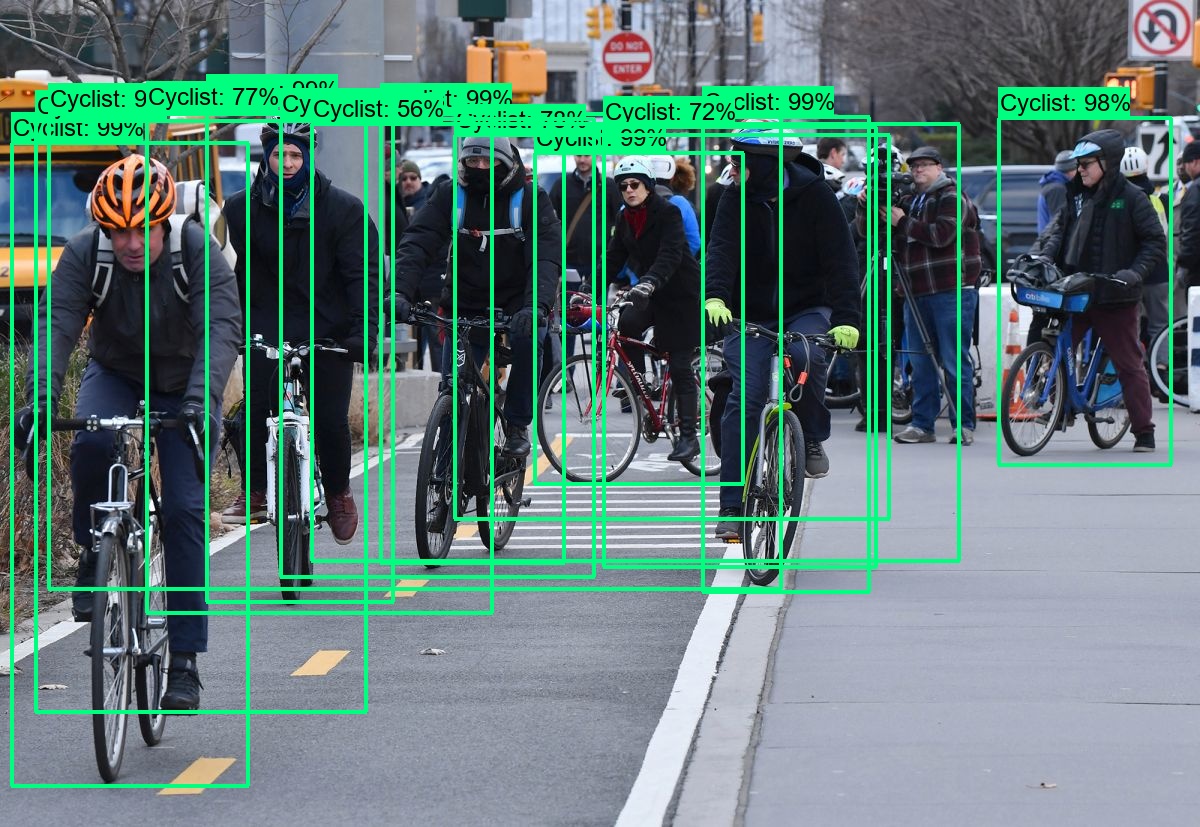} & \includegraphics[width=0.27\textwidth]{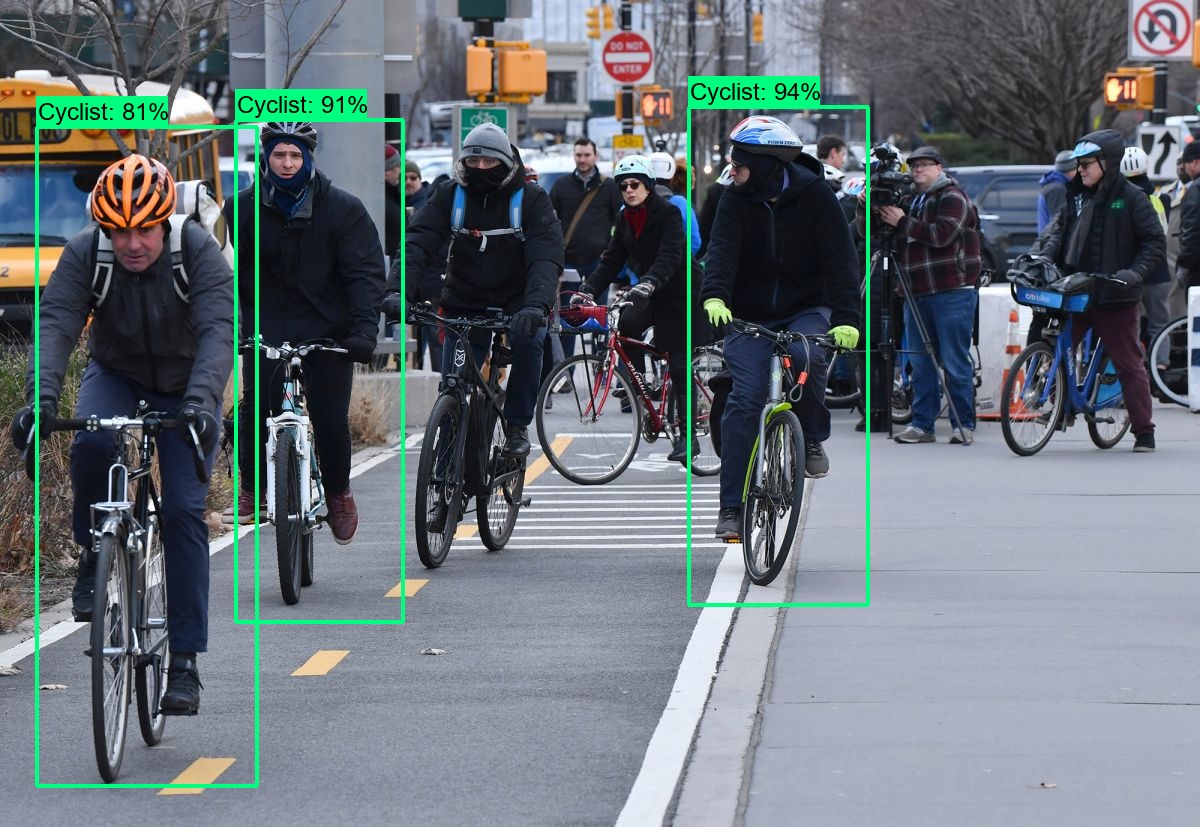} \\
 (d) FasterRCNN-InceptionResNetV2 &
(e) RFCN-ResNet101 & (f) SSD-MobilenetV2 \\[6pt]
\multicolumn{3}{c}{\includegraphics[width=0.27\textwidth]{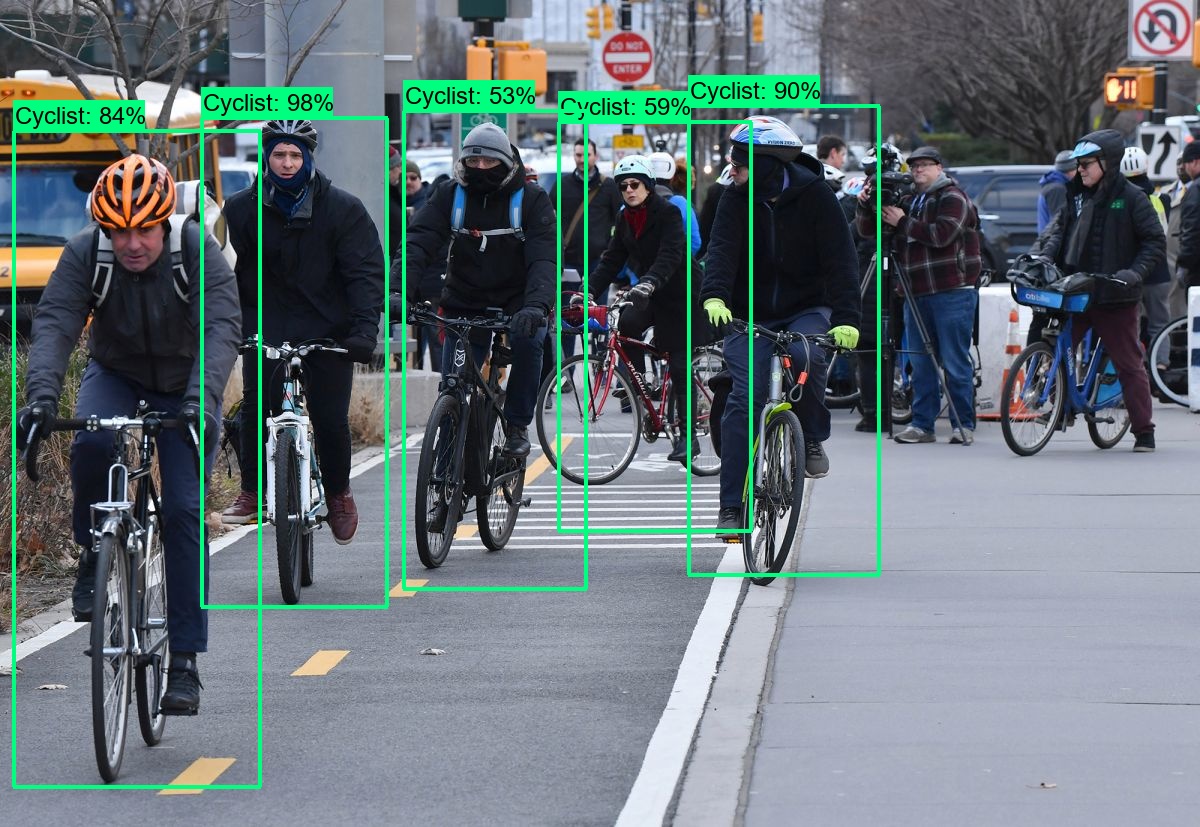} }\\
\multicolumn{3}{c}{(g) SSD-InceptionV2}
\end{tabular}
\caption{For single-class detection examples from seven different models: a) Faster R-CNN with ResNet101, (b) Faster R-CNN with InceptionV2, (c) Faster R-CNN with ResNet50, (d) Faster R-CNN with InceptionResNetV2, (e) R-FCN with ResNet101, (f) SSD with MobilenetV2 and (g) SSD with InceptionV2. The bounding box and score are display when a cyclist is detected. Most detectors showed acceptable performance. Faster R-CNN with ResNet101, Faster R-CNN InceptionV2 and Faster R-CNN with ResNet managed to identify all cyclist within an image with high score, while R-FCN with ResNet101 generated several undesired bounding boxes with different scroes for the same cyclist instances. And the other hand SSD with MobilenetV2 did not manage to detect all the cyclist, specially the ones further away.}
\label{fig:visualCyclist}
\end{center}
\end{figure*}

In this section, evaluation and implementation details of the selected meta-architectures and feature extractors are described. We provide a thoroughly comparison of the more relevant multiple object detection meta-architectures available on \cite{TensorFlow2019}, using the Detect-Bike dataset. The study is conducted in two stages, first a comparison of the performance of the main models for single class object detection is provided, for the particular case of cyclist detection in Sec. \ref{sec:CyclistDetection}. The second stage in Sec. \ref{sec:cyclistOrientationDetection}, consists in the proposal of a new multi-class detection strategy that further takes into account the orientation of the cyclists, which we consider to be of great relevance in the context of road safety of VRUs. In order to do so, we take advantage of the new dataset for cyclist detection with orientation labels Detect-Bikev2.

\begin{figure*}[ht!]
\begin{center}
\begin{tabular}{c c c}
\includegraphics[width=0.27\textwidth]{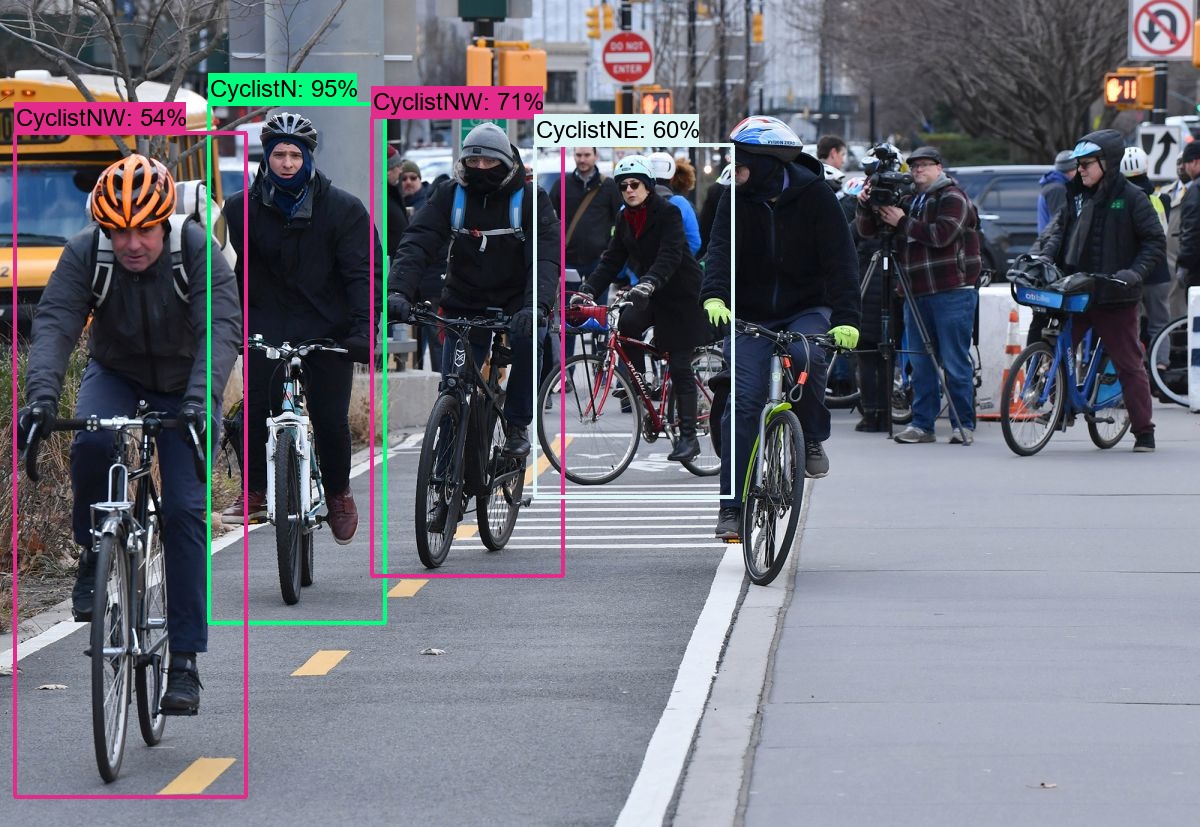} & \includegraphics[width=0.27\textwidth]{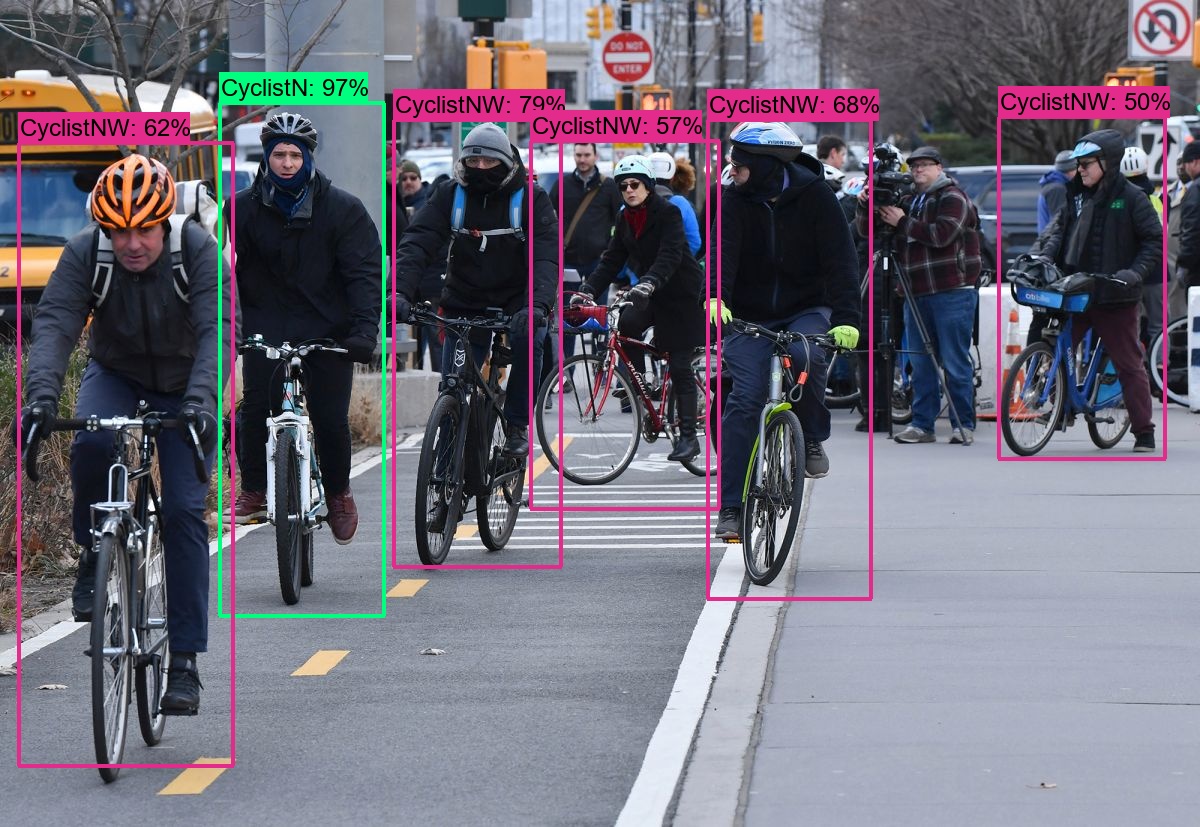} &
\includegraphics[width=0.27\textwidth]{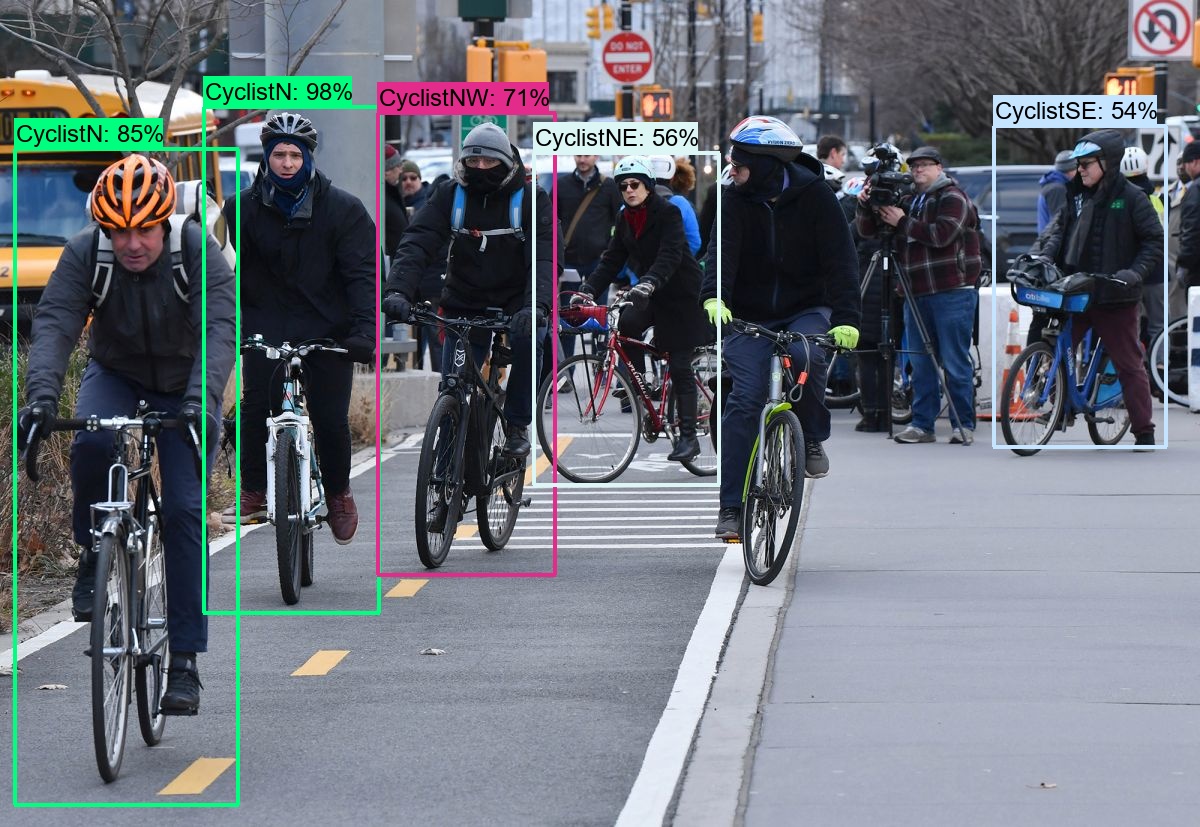} \\
(a) FasterRCNN-ResNet101 & (b) FasterRCNN-InceptionV2 & (c)  FasterRCNN-ResNet50 \\[6pt]
  \includegraphics[width=0.27\textwidth]{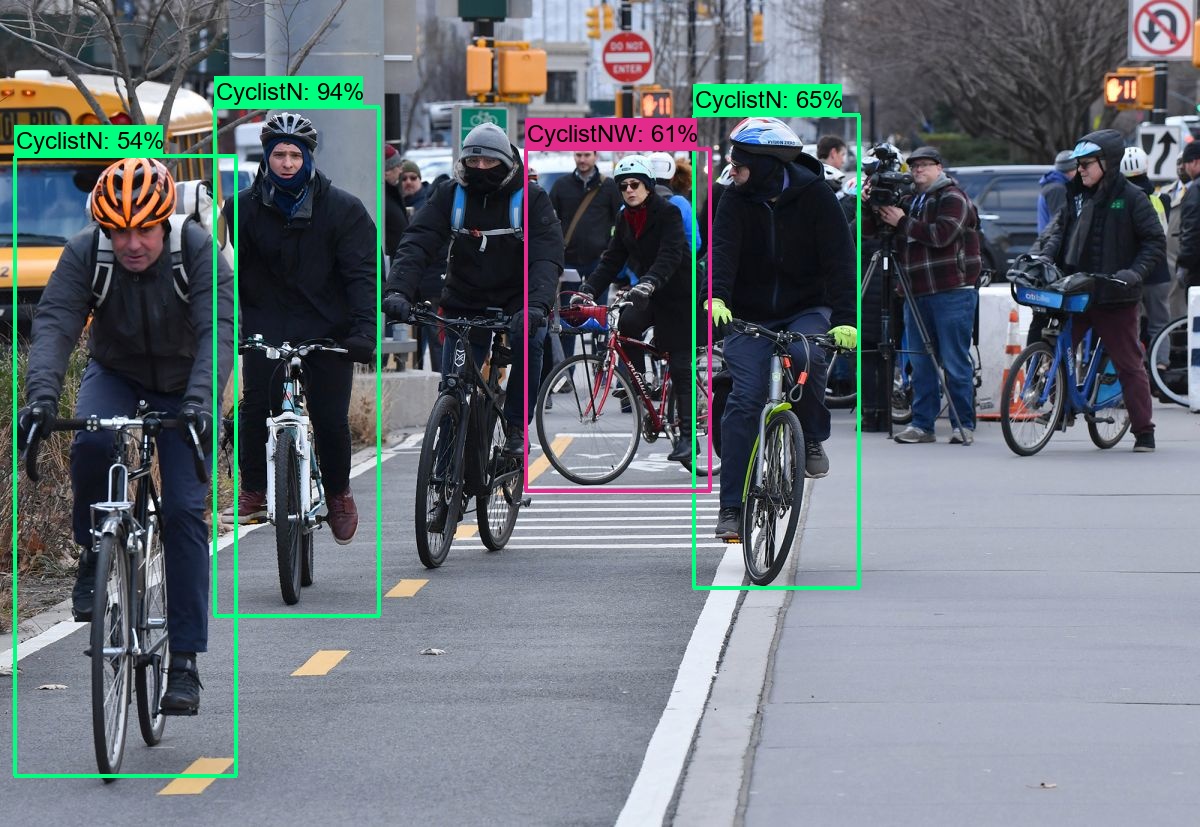} & \includegraphics[width=0.27\textwidth]{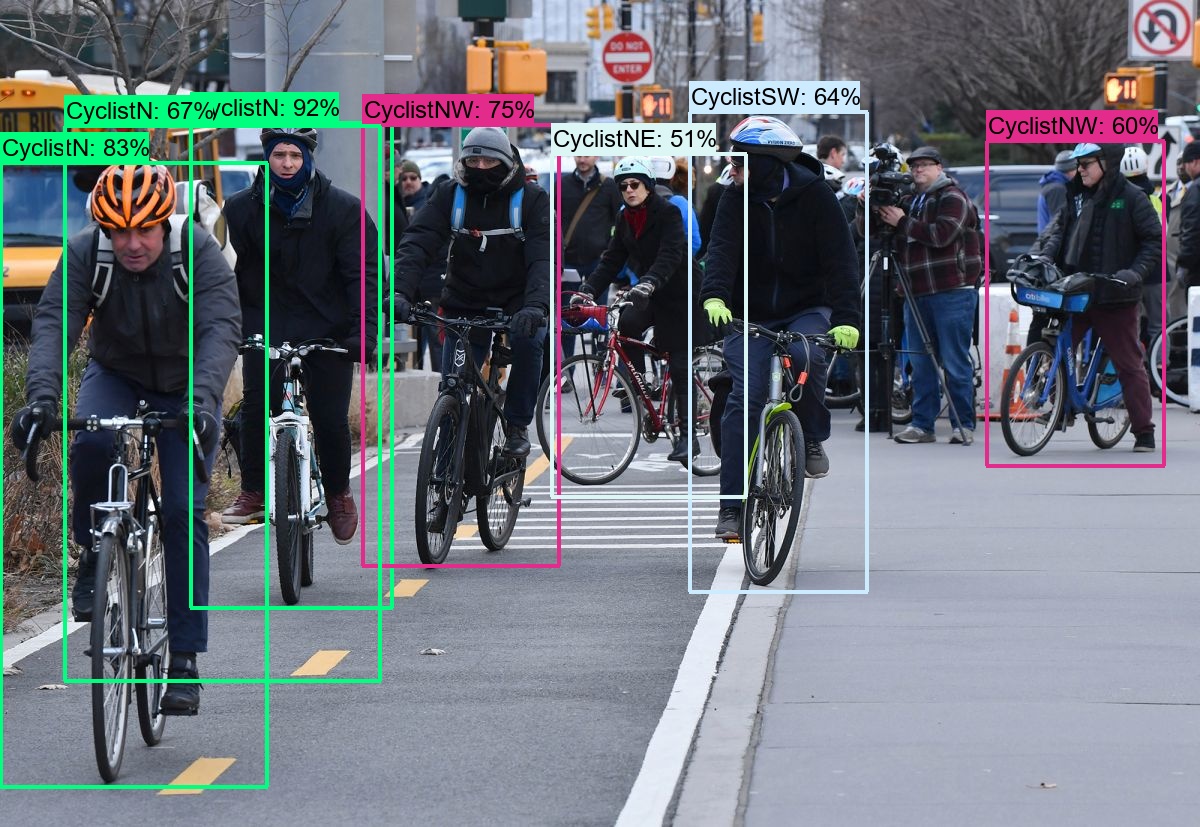} & \includegraphics[width=0.27\textwidth]{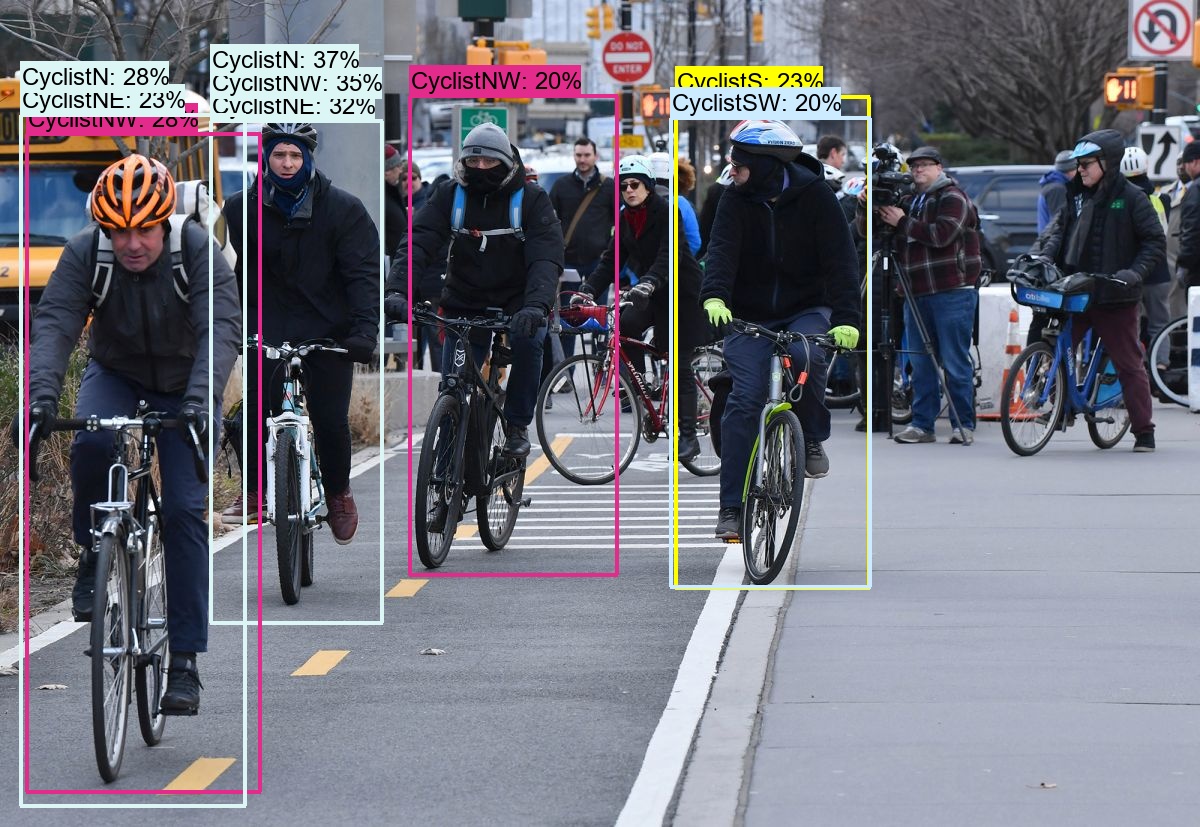} \\
  (d) FasterRCNN-InceptionResNetV2 & (e) RFCN-ResNet101 & (f) SSD-MobilenetV2 \\[6pt]
\multicolumn{3}{c}{\includegraphics[width=0.27\textwidth]{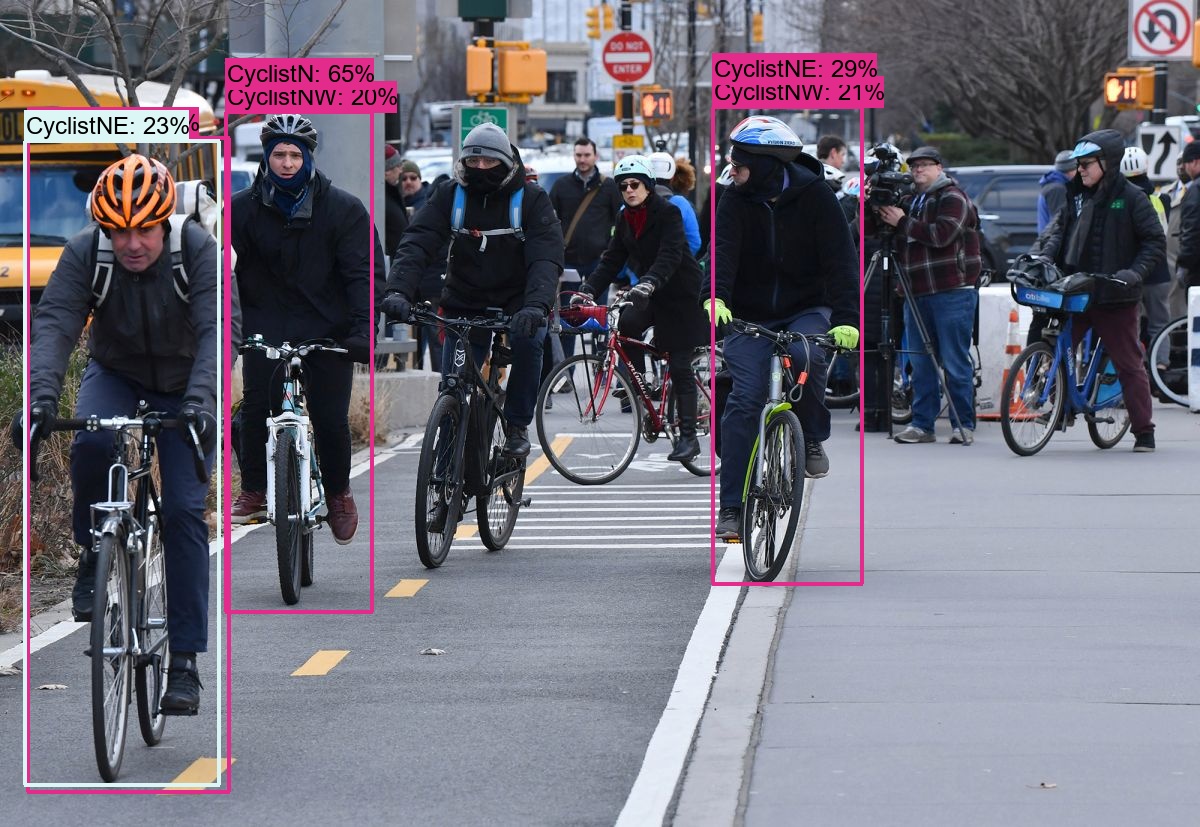} }\\
\multicolumn{3}{c}{(g) SSD-InceptionV2}
\end{tabular}
\caption{Multi-class orientation detection examples from seven different models: (a) Faster R-CNN with ResNet101, (b) Faster R-CNN with InceptionV2, (c) Faster R-CNN with ResNet50, (d) Faster R-CNN with InceptionResNetV2, (e) R-FCN with ResNet101, (f) SSD with MobilenetV2 and (g) SSD with InceptionV2. For each class a different colored bounding box is displayed. For this example, only Faster R-CNN with InceptionV2 and R-FCN with ResNet101 managed to detect all cyclists within the image. A problem that was identified is that similar classes such as CyclistNW and CyclistNE are hard to differentiate.
\label{fig:cyclistDetv3-rot}}
\end{center}
\end{figure*}

Detection examples of all models for cyclist detection with Detect-Bikev1 are shown in Fig \ref{fig:visualCyclist}, and for cyclist's orientation detection are shown in Fig. \ref{fig:cyclistDetv3-rot}. Moreover, a video demonstrating the performance of some of the different models is available at: \url{https://youtu.be/6L_MNIrCgfI}. The video is divided in two parts, the first part shows the results for single-class detection of cyclist as described in Sec. \ref{sec:CyclistDetection}, meanwhile, the second part shows the results for multi-class cyclist's orientation detection in Sec \ref{sec:cyclistOrientationDetection}. 

This work offers an updated comparison study of the state-of-the-art techniques applied to detect cyclists. We focus on identifying the best meta-architectures in terms of Average Precision (AP), execution time in Frames Per Second (FPS) and a good trade-off between both. For this evaluation, we have used Microsoft Common Objects in Context (COCO) detection Metrics. In addition, for the multi-class detection task of cyclist's orientation, we also employ the Open Images V2 detection metrics. A detailed explanation on the metrics used is provided in the following.

\subsection{Experimental setup}
\label{sub:experimentalS}

To perform the experiments, our dataset has been divided into an $80\%$ training set and $20\%$ testing set. In order to measure the speed of the detector, one video of $1920\times1080$ pixels with $435$ frames taken in the streets has been used used for obtaining the FPS required by each model using our current hardware.

The implementation has been carried out in a portable computer using Windows 10 64-bit Operative System, with a processor Intel® Core™ i7-9750H and a dedicated GPU NVIDIA® GeForce® RTX 2070 (8GB GDDR6). For the network implementation we make use of the TensorFlow-gpu V1.14.0 API, along with CUDA v10.0. Finally, the evaluation has been carried out with the help of the package python \textit{pycocotools}.

Seven models were considered and formed using a meta-architecture combined with some feature extractor, as shown in Table \ref{table:Models}. These models were trained for the detection of cyclists using the label: ``cyclist'', and the same seven models were trained but for the multi-class detection of cyclist orientation, using the $8$ classes already defined in the dataset according to orientation. 


\begin{table}[b]
\begin{minipage}[c]{0.475\textwidth}
\caption{Seven different models were generated from the combination of a meta-architecture and a feature extractor.
\label{table:Models}}
\end{minipage}
\begin{center}
\begin{tabular}{ ||l|l|| }
\hline
\multicolumn{2}{ |c| }{Models} \\
\hline
\textbf{Meta-architecture} & \textbf{Feature Extractor} \\ \hline
\multirow{3}{*} {SSD} & MobilenetV2 \\ & InceptionV2\\ \hline
\multirow{1}{*}{RFCN} & ResNet101 \\ \hline
\multirow{4}{*}{Faster RCNN} & ResNet50 \\ & InceptionV2 \\
 & ResNet101 \\ & InceptionResNetV2 \\
\hline
\end{tabular}
\end{center}
\end{table}

\subsection{Evaluation protocols}
\label{sub:evalProtocol}
For evaluation, we have used the COCO detection Metrics \cite{lin2014microsoft} and Open Images V2 detection metrics, available on \cite{Protocols}, for the comparison of each meta-architecture. COCO metrics have been selected mainly because COCO-trained models were employed for training by means of transfer learning, while Open Images V2 metrics have allowed to evaluate multi-class cyclist orientation.

In COCO, 12 metrics are handled for describing the performance of an object detector, and all of them were computed in the present study, but only the most representative ones are presented.

In order to evaluate each meta-architecture using the Detect-Bike dataset, the performance is calculated in terms of Average Precision ($AP$), which is introduced in the Pascal VOC Challenge \cite{PASCAL}. In the detection task, \textit{recall} $r$ is defined as the proportion of correct detections, or True Positives $TP$, with respect to the total number of instances given by the sum of $TP$ and False Negative $FN$ detections, where the model misses a positive detection, i.e.
\begin{equation}
  r=\frac{TP}{TP+FN}  
\label{eq:Recall}
\end{equation}
similarly, precision $p$ provides a measure on the certainty on each detection and is defined as
\begin{equation}
  p=\frac{TP}{TP+FP}
  \label{eq:Precision}
\end{equation}
where $FP$ are the False Negative detections. The $AP$ summarizes the shape of the \textit{precision/recall} curve, and is defined as the mean precision at a set of eleven equally spaced recall levels $[0,0.1,...,1]$ 
\begin{equation}
AP=\frac{1}{11}\sum_{r\epsilon\left \{ 0,0.1,...,1 \right \} }^{ }\left( P_{interp}{\left (  r\right )} \right )
\label{eq:InterpolateAP}
\end{equation}
where $P_{interp}(r)$ is an interpolation function that takes the maximum measured precision at each recall level \cite{PASCAL}. For single-class detection, there is no distinction between Average Precision (AP) and mean AP (mAP). 

Another important metric is the Intersection Over Union (IoU), which is used to obtain the area of overlap between the predicted bounding box $B_{p}$ and the ground truth bounding box $B_{gt}$
\begin{equation}
IoU=\frac{area\left ( B_{p}\cap B_{gt}  \right )}{area\left ( B_{p}\cup  B_{gt}  \right )}
\label{IoU}
\end{equation}
Then, AP can be also averaged over multiple IoU values between $0.5$ and $0.95$ thresholds, such as $AP@.50IoU$ (PASCAL VOC metric) and $AP@.75IoU$. Other scores are Average Recall (AR), which measures the maximum recall given a fixed number ($1$, $10$ or $100$) of detections allowed in the image. Both AP and AR are averaged over three instance sizes: 
\begin{itemize}
  \item Small: objects with $area< 32\ pixels^{2}$.
  \item Medium: objects with $32^{2}<area<96\ pixels^{2}$.
  \item Large: objects with $area>96\ pixels^{2}$.
\end{itemize}

The Open Images V2 metric provides the AP by category on $AP@0.5IoU$, and we used it only for evaluating multi-class detection for each meta-architecture, in the case of orientation detection. 

Also, as an important part of the functioning of each meta-architecture, loss functions are evaluated to help minimize the error in classification and localization of an object of interest. For the final classifier, classification loss $L_{cls}$ is defined as a log loss of the true class $u$ \cite{girshick2015fast}, i.e. \begin{equation} L_{cls}=-log(p_u)\end{equation}
where $p_u$ is the probability distribution for class $u$. On the other hand, the localization Loss $L_{loc}$ represents the error of the bounding box regressor and is defined as
\begin{equation}
	L_{loc}\left ( t^{u}, v \right ) = \sum_{i \in \left \{ x,y,w,h \right \}}^{ } smooth_{^{L_{1}}} \left ( t^{u}_i - v_{i} \right )
    \label{loss_loc}
\end{equation} 
where the $smooth_{L1}$ loss is used, it is,  for a variable $\xi$
\begin{equation}
    smooth_{L_1} \left ( \xi \right ) = \begin{cases}
    0.5\xi^{2} & if\left | \xi \right | <  1\\ 
    \left | \xi \right | - 0.5 &  otherwise,
    \end{cases}
    \label{smoothL}
\end{equation}
there, a ground truth for class $u$ and a ground-truth bounding box regression for target $v$, the $L_{loc}$ is defined over a four-tuple $(v_x, v_y, v_h, v_w)$ for top-left corner $(x,y)$ and height and width $(h,w)$, and a predicted tuple $t^{u}=\left ( t_{x}^{u}, t_{y}^{u}, t_{w}^{u},t_{h}^{u}\right )$ for the ground-truth class $u$. In addition, for region proposal based meta-architectures with stage of Region Proposal Network, the objectness loss, that indicates if a bounding box is an object or part of the background, is also considered \cite{fasterObject2015}.

Other important aspect to evaluate the performance of the detection algorithms is the execution speed in Frames Per Second (FPS) \cite{liu2020deep}. In this sense, we calculated the mean Frames Per Second for each detection model using the same hardware on a video of $1920\times1080$ with $435$ frames.

\subsection{Cyclist Detection }
\label{sec:CyclistDetection}

In this sub-section, we focus on the single-class cyclist detection, in order to identify which state-of-the-art techniques are the best suited for detection of these particular case of VRU.

\begin{figure}[t]
\begin{center}
  \subfloat[][Average Precision by size.]{
   \label{fig:Cyclist_AP}
    \includegraphics[width=0.44\textwidth]{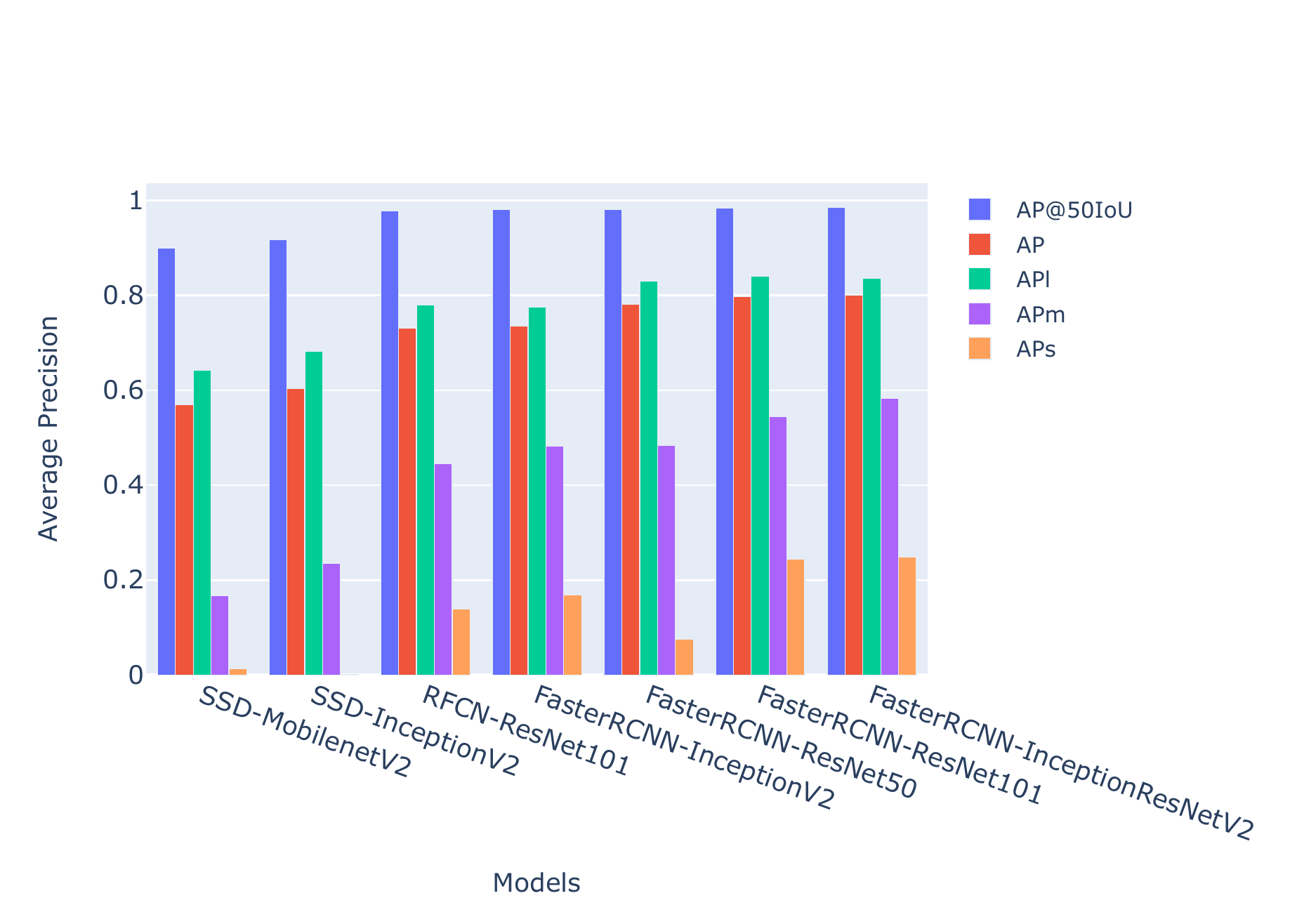}}
    \newline
    \vspace{-0.3cm}
  \subfloat[][Classification Loss]{
   \label{fig:Cyclist_Loss}
    \includegraphics[width=0.42\textwidth]{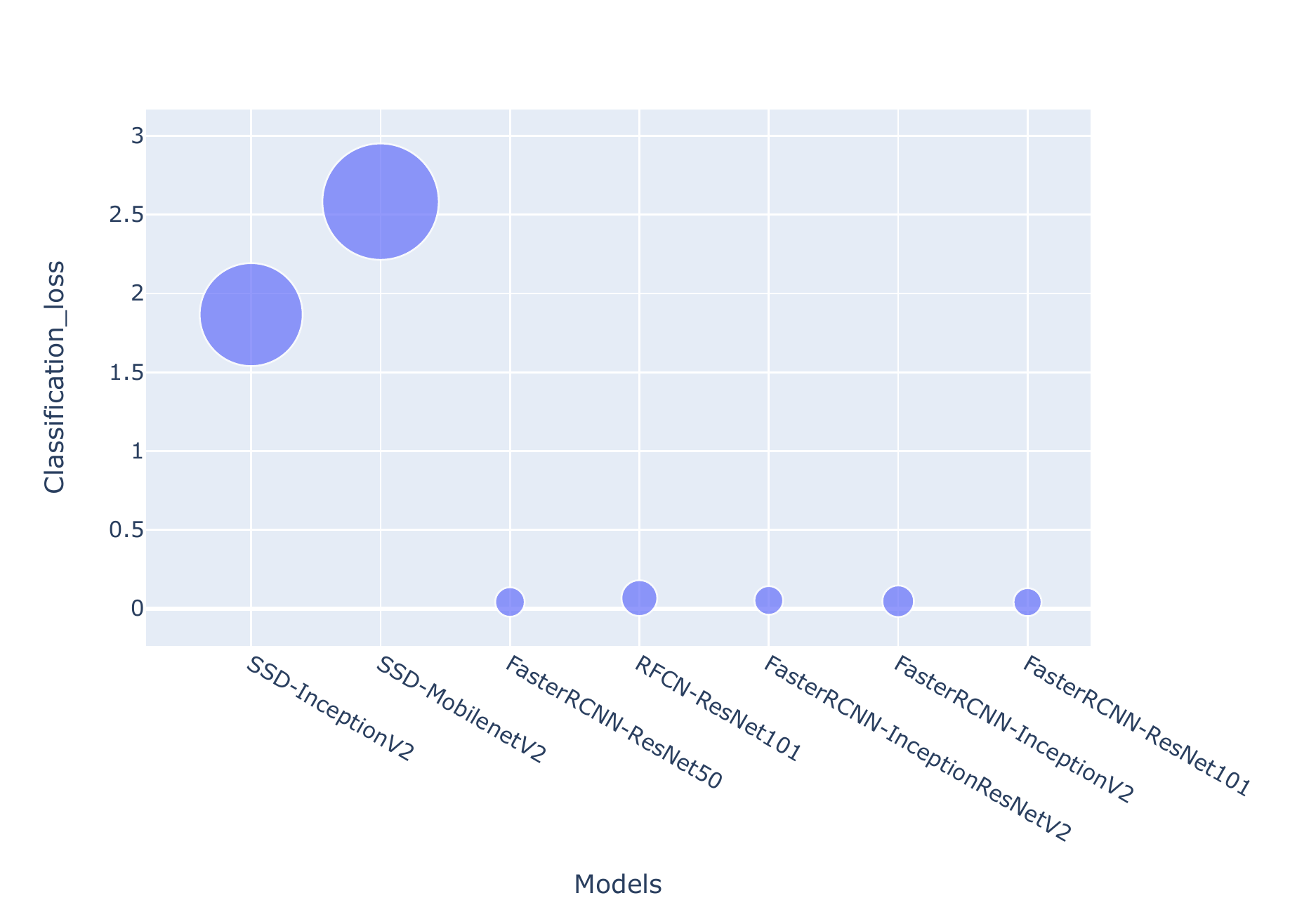}}
 \caption{Single-class cyclist detection using Detect-Bikev1 a) Average Precision with threshold 0.5 and 0.7 on IoU. All models based on the Faster R-CNN meta-architecture and R-FCN achieved similar results for the cyclist detection task, from which Faster R-CNN with InceptionResNetV2 was the most precise.  b) Classification loss (axis-y) and localization loss (size) for each model. 
 \label{fig:Cyclist_Results}}
\end{center}
\end{figure}

The seven models that have been selected by combining a meta-architecture with a specific feature extractor, shown in Table \ref{table:Models}, have been trained for the instance ``cyclist'', using the dataset Detect-Bikev1, and evaluated with the metrics explained in Sec. \ref{sub:evalProtocol}. Even though all the metrics proposed by COCO \cite{lin2014microsoft} have been analyzed in this work, only the most relevant ones are presented through Figures \ref{fig:Cyclist_AP} and  \ref{fig:Cyclist_Loss}, along with Tables \ref{table:Cyclist_AP} and \ref{table:Cyclist_Loss}.

Considering the size of the objects is important for the cyclist's detection on the roads, since it considerably affects the detection performance (further away objects are smaller and harder to detect), moreover, it is a good indicator of how close cyclists are to the camera, where larger objects present  higher collision risks. In Fig. \ref{fig:Cyclist_AP} the AP performance for each of the models evaluated by object size (large, medium and small) is presented, along with the $AP@50IoU$ and the overall AP. From there, it can be observed how their performance changes with respect to the size of the cyclist instances, Faster meta-architecture R-CNN and R-FCN were the more efficient if we consider the medium and small instances as opposed to SSD. Further information is included in Table  \ref{table:Cyclist_AP}, with the most important metrics to evaluate the good functioning of the each model. 

For cyclist's detection, Faster R-CNN meta-architecture using InceptionResNetV2 feature extractor was found to be the most precise, with the highest AP for medium and small cyclist sizes, $AP_{m}$ and $AP_{s}$ respectively. Meanwhile, Faster R-CNN using ResNet101 feature extractor achieved betters result for AP large $AP_{l}$, as well as for $AP@.50IoU$ and $AP@.75IoU$. Also, for AR in all cyclist sizes, again Faster R-CNN meta-architecture was in the top detectors, as can be appreciated in Table \ref{table:Cyclist_AP} and Fig. \ref{fig:Cyclist_AP}. These results are consistent with the ones reported in the literature for object detection. Nevertheless, it is important to notice that there is a trade-off between precision and detection time, where the most precise algorithm FasterRCNN with InceptionResNetV2 is up to $30$ times slower than the fastest model, SSD with MobilenetV2, when running on our baseline hardware. This stresses the importance on selecting the most suitable technique according to the particular scenario. 

When considering the AP with threshold 0.50 by IoU, all models showed competitive results to address the cyclist's detection. Regarding Faster R-CNN with 50-layers and 101-layers it was observed that indeed there is no significant increase in the values of AP obtained, even with 50-layers a better AR was obtained for all sizes of cyclist, along with a faster response. On the other hand, R-FCN with ResNet101 obtains very similar results to Faster R-CNN with InceptionV2, which indicates that placing a more robust feature extractor benefits the meta-architecture classifier. Besides, when considering the time response, Faster R-CNN with InceptionV2 was superior.

Nonetheless, SSD was found to be far superior in time response with respect to the other meta-architectures, and despite having problems with medium and small cyclists, it has achieved good results for large size cyclists, specially when combined with the InceptionV2 feature extractor. Hence, it appears as a suitable alternative for real-time applications where far away cyclists can be neglected in order to get a faster response.

On the other side, it is important to analyze the results in the detection loss, which are presented in Table \ref{table:Cyclist_Loss}, since these assessments are relevant because they provide information about how well the bounding box covers the cyclist (localization loss) and how well the class is detected (classification loss). From there, Faster R-CNN using ResNet101 obtained the lowest losses in both the classification and location of the box classifier, and also in terms of localization and objectness loss of the RPN. This shows that increasing the architecture complexity, allows better results in both classification loss and localization loss. Nevertheless, when evaluating time response, Faster R-CNN and R-FCN meta-architectures are strongly overcome by the SSD meta-architecture, as can be appreciated in Table \ref{table:Cyclist_Loss}. This is why it is important to select the best model according to the application, and determine good trade-offs between the parameters of interest. In particular, for the case of  VRU safety on the road with ITS, it is important to detect them correctly, but it is also relevant to detect them on time. Furthermore, having detections at high rate can be of great use, specially when combined with other algorithms, for example to track the objects in real-time.

\begin{table*}[hb!]
\begin{minipage}[c]{0.95\textwidth}
\caption{COCO metrics for each model, AP@.5IoU, AP@.95IoU, AP and AR for all sizes of cyclist's instances and FPS. All models exceed 89\% AP@.5IoU. Faster R-CNN meta-architecture with InceptionResNetV2 feature extractor was the most precise and SDD meta-architecture with MobilenetV2 feature extractor was the speediest, while Faster R-CNN meta-architecture with InceptionV2 offers a good trade-off between precision and time response, but SSD meta-architecture with InceptionV2 feature extractor was the best choice for real-time applications, specially if far away objects are neglected.
\label{table:Cyclist_AP}}
\end{minipage}
\addtolength{\tabcolsep}{-1pt}
\begin{center}
\begin{tabular}{|rrrrrrrrrrrrr|}
\toprule
AP@.5 &  AP@.75 &       AP &      $AP_{l}$ &      $AP_{m}$ &      $AP_{s}$ &  AR& $AR_{l}$ &    $AR_{m}$ &   $AR_{s}$  & FPS & Architecture &  {Feature Extractor}\\
\midrule
\midrule
0.899 &  0.670 &  0.5691 &  0.6426 &  0.1673 &  0.0129 &  0.6365 &  0.7087 &  0.3005 &  0.0125 &  \textbf{47.1278}&          \textbf{SSD} &        MobilenetV2 \\
\textbf{0.918} &  0.699 &  0.6042 &  0.6820 &  0.2348 &  0.0018 &  0.6690 &  0.7349 &  0.3642 &  0.0250 &  \textbf{33.7189} &          \textbf{SSD} &        \textbf{InceptionV2} \\
0.978 &  0.863 &  0.7316 &  0.7795 &  0.4457 &  0.1391 &  0.7706 &  0.8135 &  0.5732 &  0.2750 &   7.7476 &         RFCN &          ResNet101 \\
\textbf{0.981} &  \textbf{0.882} &  \textbf{0.7354} &  0.7757 &  0.4822&  0.1692 &  0.7766& 0.8156 & 0.6005 &  0.1750&  \textbf{8.9099} &   \textbf{FasterRCNN} &        \textbf{InceptionV2} \\
0.981 &  0.891 &  0.7808 &  0.8303 &  0.4831 &  0.0757 &  0.8139 &  0.8592 &  0.6050 &  0.3125 &  6.4775 &   FasterRCNN &           ResNet50 \\
0.984 &  0.910 &  0.7973 &  \textbf{0.8410} &  0.5445 &  0.2440 &  0.8275 &  0.8674 &  0.6442 &  \textbf{0.3625} &   5.1536 &   FasterRCNN &          ResNet101 \\
\textbf{0.985} &  \textbf{0.936} &  \textbf{0.8007} &  0.8364 &  \textbf{0.5827} &  \textbf{0.2488} &  \textbf{0.8352} &  \textbf{0.8685} &  \textbf{0.6861} &  0.2625 &   1.5618 &   \textbf{FasterRCNN} &  InceptionResNetV2 \\
\bottomrule
\end{tabular}
\end{center}
\end{table*}

\begin{table*}[hb!]
\begin{minipage}[c]{0.95\textwidth}
\caption{For all the considered models, the classification loss of bounding box (Classification-Box) and localization loss of bounding box (Localization-Box) were obtained, and in the same way, the localization loss (Localization-RPN) and objectness loss (Objectness-RPN) of the RPN-based methods are presented. Note that regression/classification based methods such as SSD do not provide this losses. Here, Faster R-CNN  and R-FCN architectures were characterized by low values in all metric losses.
\label{table:Cyclist_Loss}}
\end{minipage}
\addtolength{\tabcolsep}{-1pt}
\begin{center}
\begin{tabular}{llllll}
\toprule
{Classification-Box} &  {Localization-Box}&  {Localization-RPN} & {Objectness-RPN} & Architecture &  Feature\_Extractor \\
\midrule
\textbf{0.041877} &      \textbf{0.025038} &  \textbf{0.094797} &   0.070329 &   \textbf{FasterRCNN} &          \textbf{ResNet101} \\
 0.042922 &      0.027770 &  0.097316 &   \textbf{0.051907} &   \textbf{FasterRCNN} &\textbf{ResNet50} \\
0.047551 &      0.031910 &  0.094501 &   0.052287 &   FasterRCNN &        InceptionV2 \\
0.052672 &      0.026272 &  0.153935 &   0.103883 &   FasterRCNN &  InceptionResNetV2 \\
0.067570 &      0.040941 &  0.107662 &   0.068421 &         RFCN &          ResNet101 \\
1.865798 &      0.337980 &         - &          - &          SSD &        InceptionV2 \\
2.582322 &      0.431545 &         - &          - &          SSD &        MobilenetV2 \\
\bottomrule
\end{tabular}
\end{center}
\end{table*}

\subsection{Cyclist Orientation Detection}
\label{sec:cyclistOrientationDetection}

Cyclist detection is not a trivial task, however, nowadays it has become a much more challenging problem when considering the cyclist's orientation, reason why the focus of this work is on identifying the cyclist's direction of movement based on his orientation. Once we have evaluated the meta-architectures along with the main feature extractors using Detect-Bikev1, in this sub-section we present the evaluation for multi-class detection, where each class is considered one of the cyclist's orientations.

\begin{figure}[ht!]
\begin{center}
  \subfloat[][Average Precision (AP) and Average Recall (AR) by large size]{
   \label{fig:Cyclist_Orientation_mAP}
    \includegraphics[width=0.44\textwidth]{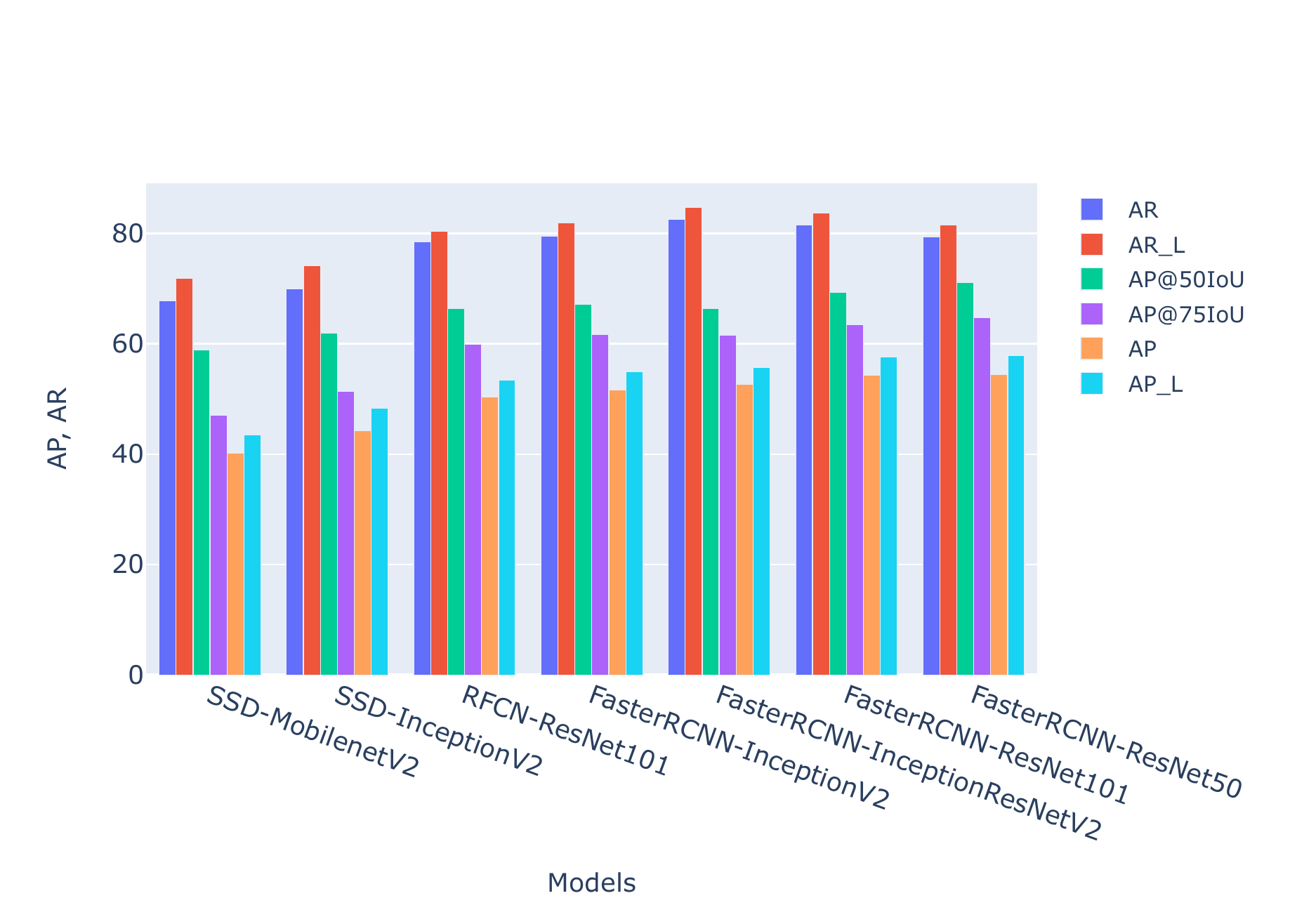}}
    \newline
  \subfloat[][AP by cyclist's orientation class]{
   \label{fig:Cyclist_Orientation_ByClass}
    \includegraphics[width=0.44\textwidth]{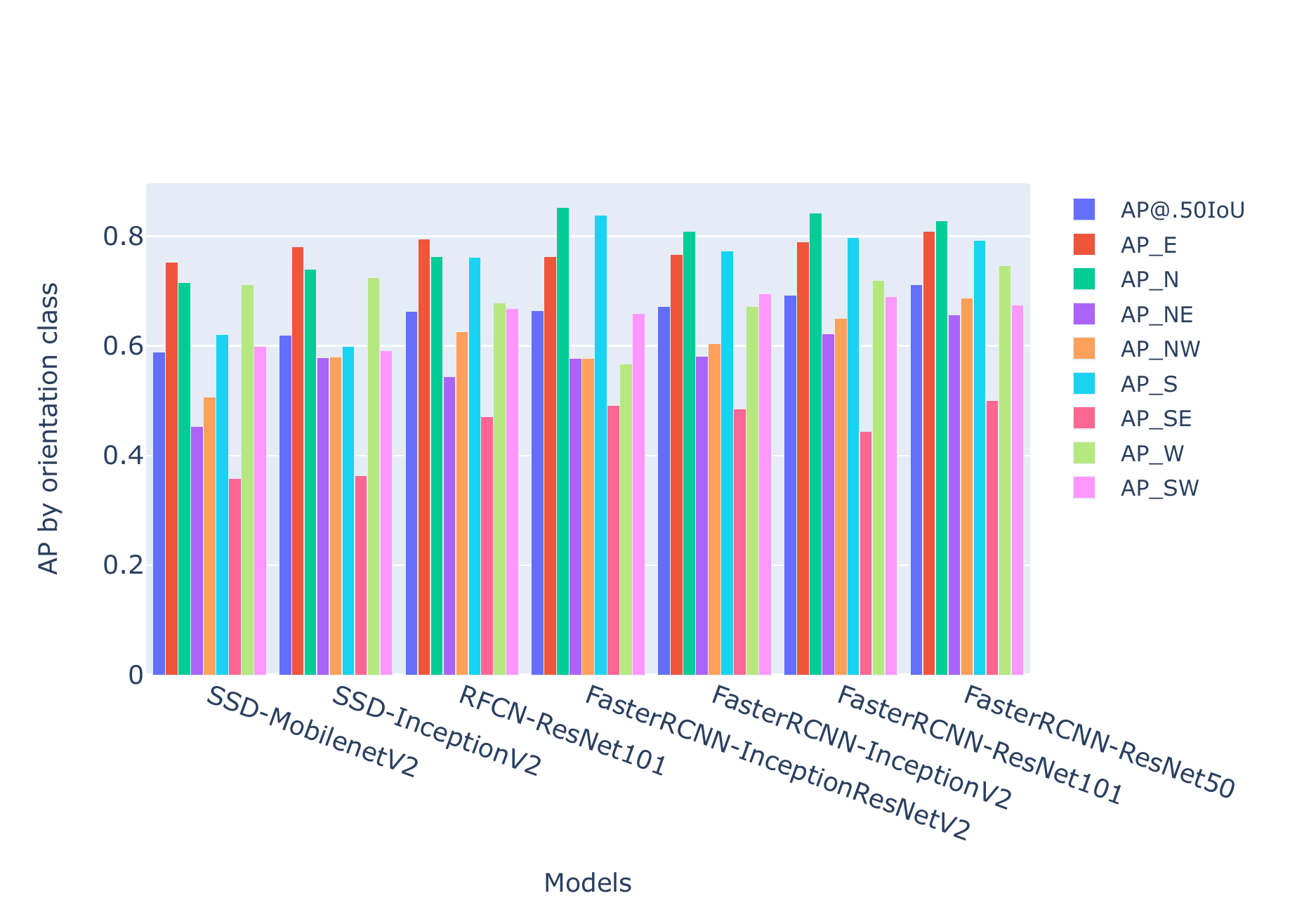}}
 \caption{Cyclist orientation detection a) mean Average Precision  (mAP) with threshold 0.5 and 0.7 on IoU for large size. b) Average Precision with threshold 0.5 on IoU for each class. Faster R-CNN with ResNet50 was the most consistent meta-architecture for all classes. In all the models it is observed that the number of instances by class considerably affects the detection performance.
 \label{fig:Cyclist_Orientation_Results}}
\end{center}
\end{figure}

\begin{figure}[ht!]
\begin{center}
  \includegraphics[width=0.39\textwidth]{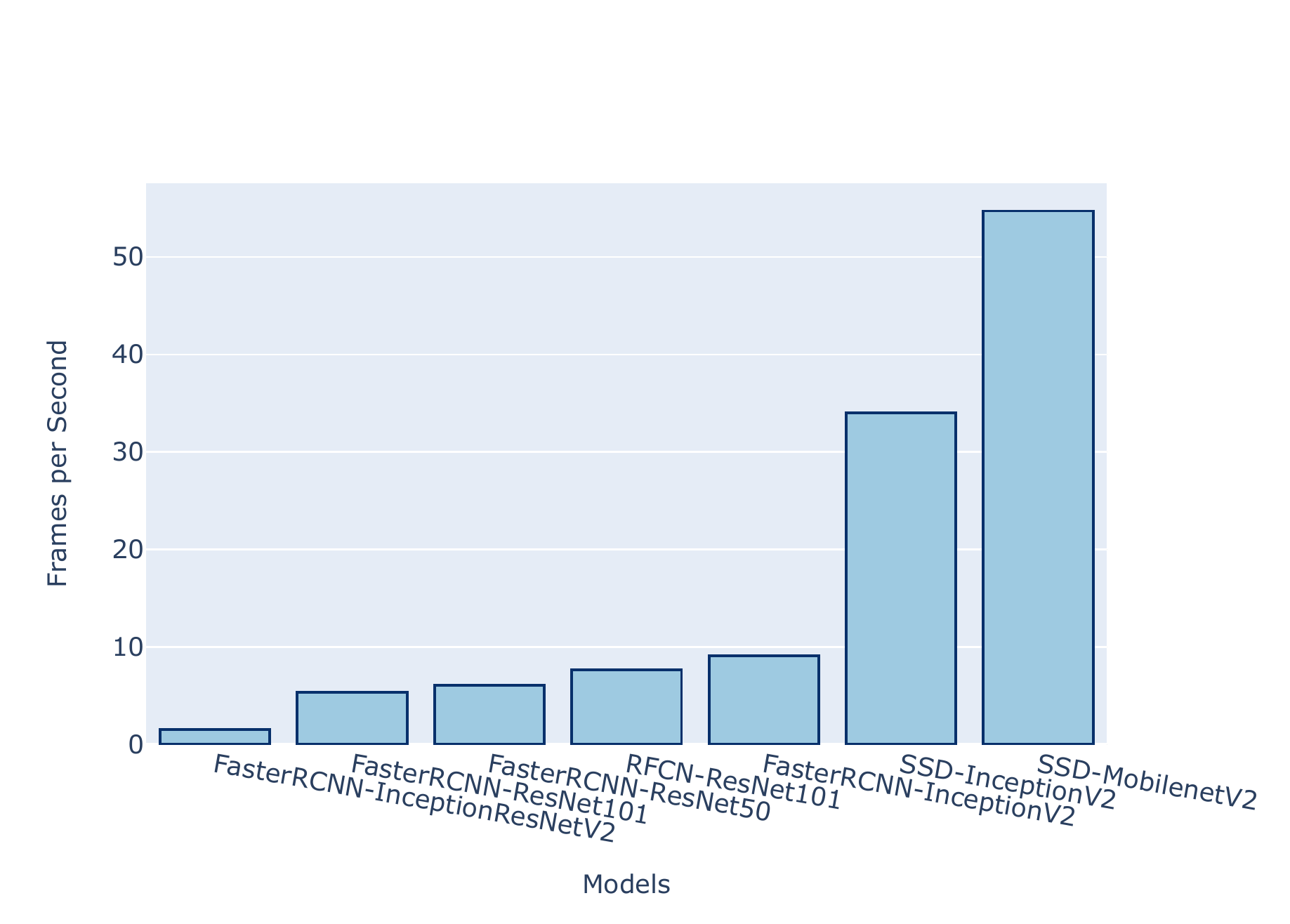}
\caption{Execution time in FPS for each model measured using the same video on a baseline hardware. SSD-MobilenetV2 is the fastest model with $54.71 FPS$, considerably overcoming the region-based models. Meanwhile, among the  region-based strategies, FasterRCNN-InceptionV2 turned out to be fastest with $9.1FPS$.}
\label{fig:Cyclist_Orientation_FPS}       
\end{center}
\end{figure}

In this case we have provided and used for training the Detect-Bikev2 dataset, with the cyclist's orientation labels, and evaluated with the metrics explained in Sec. \ref{sub:evalProtocol}. Analogous to the previous subsection \ref{sec:CyclistDetection}, the most relevant metrics are presented in Fig. \ref{fig:Cyclist_Orientation_mAP}, and for time response the FPS in Fig. \ref{fig:Cyclist_Orientation_FPS}, along with Table \ref{table:Cyclist_Orientation_mAP}. Also, since the orientation detection problem is accomplished as a multi-class detection, with eight different classes, we further employ the Open Image V2 metrics such as AP by category of cyclist orientation on AP@.50IoU, as depicted in Fig. \ref{fig:Cyclist_Orientation_ByClass} along with Table \ref{table:Cyclist_Orientation_ByClass}.

It is noteworthy to point out that the new introduced labeled dataset Detect-Bikev2 is still under construction, and the number of instances by class are not perfectly balanced, containing mostly large instances. In this study we focus mainly on large instances, since as an starting point, large instances are the more critical to avoid collisions. 
The majority of models trained with this new labeled dataset were competitive to perform the cyclist orientation detection, as can be appreciated in Fig. \ref{fig:Cyclist_Orientation_mAP}, except for the class CyclistSE, which fails mainly because the detectors frequently confuse it with CyclistSW. In order to correct this issue, more instances of these two classes must be added to the dataset, however, it is worth mentioning that such confusion is not critical for the considered scenario, provided that both classes are very similar and the cyclist is going away from the camera. Also, further employing tracking techniques would considerably mitigate such problem.

For cyclist's orientation detection, in general Faster R-CNN meta-architecture with ResNet50 feature extractor obtained the best results for this multi-class detection. Surprisingly, ResNet with 50-layers outperformed precision-wise ResNet with 101-layers. This suggests that deeper networks require more instances to work better. Besides, all models with Faster R-CNN meta-architecture managed good result for this task. 


We can observe that the SSD meta-architecture works correctly for large instances. On the other side, for all meta-architectures, we note that they are significantly degraded in precision performance  by classes similar in appearance, particularly CyclistNE vs CyclistNW and CyclistSE vs CyclistSW. (see Tables \ref{table:CyclistDBOrientation} and \ref{table:Cyclist_Orientation_ByClass}). This issue can be corrected by adding more instances of these classes in the database.


Also, Fig. \ref{fig:Cyclist_Orientation_FPS} presents the time response in FPS, using a testing video on our baseline hardware, for each model, where it can be identified that despite implementing multi-class detection, the response times are comparable with those obtained for single-class detection in Section \ref{sec:CyclistDetection}. This Figure suggests that for scenarios where time response is important for orientation detection of mainly large cyclists, the best option is SSD with InceptionV2.



In summary, for cyclist's orientation detection, the evaluation suggests that Faster R-CNN meta-architecture with InceptionV2 feature extractor allows for a good trade-off between precision and time response, offering a good performance considering AP, AR, localization and classification loss, as stated in Table \ref{table:Cyclist_Orientation_ByClass}. Moreover, in terms of response time for the region-based methods considered in this study, FasterRCNN-InceptionV2 was the fastest. On the other hand, if the main objective is to obtain a fast model for real-time applications, and high speed is required, the best trade-off is obtained with SSD meta-architecture using InceptionV2 feature extractor. This is a good strategy to detect the orientation of the cyclist, even if it is not the most precise, since it allows to obtain the notion of the cyclist's movement at high rate, and is suitable to be implemented embedded on a low-cost vehicle with limited computation. Furthermore, this can be improved when combined with tracking techniques, which usually require fast estimation updates of the cyclist's position and orientation. In addition, this model achieves considerably faster detections compared to other strategies reported in the literature for orientation detection \cite{guindel2018fast,guindel2019traffic}.

\begin{table*}[hb!]
\begin{minipage}[c]{0.95\textwidth}
\caption{COCO metrics for each model, AP@.5IoU, AP@.95IoU, AP and AR for large and medium sizes of cyclist instances, classification loss ($cl_{loss}$), localization loss ($loc_{loss}$) and FPS. Faster R-CNN meta-architecture with ResNet50 feature extractor was the most precise, while SDD meta-architecture with MobilenetV2 feature extractor was the speediest.
\label{table:Cyclist_Orientation_mAP}}
\end{minipage}
\addtolength{\tabcolsep}{-1pt}
\begin{center}
\begin{tabular}{|lrrrrrrrrrrrl|}
\toprule
AP@.5 &  AP@.75 &     mAP &   $AP_{l}$ &  $AP_{m}$  &      AR &    $AR_{l}$  &    $AR_{m}$  &  $cl_{loss}$ & $loc_{loss}$ &    FPS & Architecture &  Feature\_Extractor \\
\toprule
\midrule
0.5890 &  0.471 &  0.4022 &  0.4354 &  0.0619 &  0.6775 &  0.7183 &  0.3199 &     3.1016 &    0.3290 &  \textbf{54.7106} &          SSD &        MobilenetV2 \\
0.6190 &  0.514 &  0.4426 &  0.4836 &  0.0496 &  0.6992 &  0.7412 &  0.2931 &     2.9726 &    0.2906 &  34.0071 &          SSD &        InceptionV2 \\
0.6634 &  0.599 &  0.5035 &  0.5348 &  0.1628 &  0.7845 &  0.8042 &  0.5725 &     0.1661 &    0.0491 &   7.6540 &         RFCN &          ResNet101 \\
0.6720 &  0.617 &  0.5170 &  0.5496 &  0.1693 &  0.7951 &  0.8198 &  0.5899 &     0.1337 &    0.0353 &   9.1028 &   FasterRCNN &        InceptionV2 \\
0.6640 &  0.616 &  0.5262 &  0.5572 &  \textbf{0.1988} &  \textbf{0.8259}&  \textbf{0.8469} &  0.6095 &     0.1305 &    \textbf{0.0295} &   1.5290 &   FasterRCNN &  InceptionResNetV2 \\
0.6930 &  0.635 &  0.5438 &  0.5763 &  0.1950 &  0.8155 &  0.8376 &  \textbf{0.6122} &     0.1365 &    0.0306 &   5.3274 &   FasterRCNN &          ResNet101 \\
\textbf{0.7117} & \textbf{ 0.647} &  \textbf{0.5448} &  \textbf{0.5782} &  0.1829 &  0.7939 &  0.8160 &  0.5996 &     \textbf{0.1166} &    0.0350 &   6.0998 &   FasterRCNN &           ResNet50 \\
\bottomrule
\end{tabular}
\end{center}
\end{table*}

\begin{table*}[hb]
\begin{minipage}[c]{0.95\textwidth}
\caption{Average Precision with Threshold AP@.50 by IoU using the Detect-Bikev2 dataset for cyclist orientation detection. The three most efficient meta-architectures were Faster R-CNN with ResNet50, Faster R-CNN with InceptionResNetV2, and Faster R-CNN with ResNet101, but also SSD with InceptionV2 managed to obtain relatively good results in most classes, which positions it as a good alternative in orientation detection.
\label{table:Cyclist_Orientation_ByClass}}
\end{minipage}
\addtolength{\tabcolsep}{-1pt}
\begin{center}
\begin{tabular}{lllllll}
\toprule
\textbf{CyclistE} & \textbf{CyclistN}&\textbf{CyclistNE} & \textbf{CyclistNW} & \textbf{Architecture} &  \textbf{Feature Extractor}\\
\midrule
0.753154		 &  0.716062	 &  0.453501&  0.506868		 &          SSD &          MobilenetV2 \\
0.789926	 & 0.842027	 &  0.622228			&  0.649831	 &   FasterRCNN &          ResNet101 \\
0.767118 &  0.809508&  0.581257		 &  0.604005		 &   FasterRCNN &        InceptionV2 \\
0.781269 &  0.739766	 &  0.579011	 &  0.579609		 &          SSD &        InceptionV2 \\
0.794934 &  0.763203	 &  0.544258	 &  0.625844 &         RFCN & ResNet101\\
0.763053	 &  \textbf{0.852115} &  0.577617	 &  0.577302	 &  \textbf{FasterRCNN} &  \textbf{InceptionResNetV2} \\
\textbf{0.808598}	 &  0.828415	 & \textbf{0.657107}	& \textbf{0.686987}		 &  \textbf{FasterRCNN} &  \textbf{ResNet50} \\
\bottomrule
\toprule
\textbf{CyclistS }& \textbf{CyclistSE} & \textbf{CyclistW} & \textbf{CyclistSW} & \textbf{Architecture} &  \textbf{Feature Extractor} \\
\midrule
0.620308 &  0.358924	 & 0.711381 & 0.599037 &          SSD &          Mobilenetv2 \\
\textbf{0.797039}		 & 0.443562 &  0.719340	 &  0.690445	&   FasterRCNN &          ResNet101 \\
0.773338	 & 0.484934	 &  0.672358	& \textbf{0.695525} &   \textbf{FasterRCNN} &\textbf{InceptionV2} \\
0.599015		 & 0.363876	& 0.724558	 &  0.591662 &      \textbf{SSD} &\textbf{InceptionV2} \\
0.761743 &  0.470963 &  0.678240 &  0.668205	 &         RFCN &          ResNet101 \\
0.838598		 &0.491344	& 0.567436	&  0.658789	 &FasterRCNN&InceptionResNetV2 \\
0.791745	 & \textbf{0.499857}	& \textbf{0.746269}		 &  0.674663 &\textbf{FasterRCNN}&\textbf{ResNet50} \\
\bottomrule
\end{tabular}
\end{center}
\end{table*} 

\section{Conclusion and Future Work}
\label{sec:Conclusion}

In this work, we propose a new approach to improve the safety on the road of a particularly vulnerable kind of VRU in the context of ITS, where it is not enough to detect an object to prevent potential accidents, but it is also critical to predict the movement of the object in the near future. The key idea relies on the fact that two-wheeled vehicles, such as bicycles and motorbikes, always move in the forward direction, hence, knowing their orientation provides useful information about their motion. Then, we propose a multi-class object detection technique based on the state-of-the art CNN meta-architectures and feature extractors, where in addition to only detecting the object and its position, the detector also provides its orientation. It is important to note that the proposed strategy can be easily extended to other kinds of vehicles such as motorcycles or car-like vehicles.

In order to accomplish the multi-class orientation detection, we provide a new cyclist image dataset ``Detect-Bike'', which contains $20,229$ cyclist instances over $11,103$ images, labeled according to their orientation. 

Besides, we extensively compare the state-of-the-art meta-architectures SSD, Faster R-CNN and R-FCN, combined with MobilenetV2, InceptionV2, ResNet50, ResNet101 and InceptionResNetV2 feature extractors for cyclist's and their orientation detection. With this we provide an updated and broader study of the state-of-the-art methods for cyclists detection and their orientation, considering the two frameworks used for generic object detection, ``region proposal based'' and ``regression/classification based'', and analyzing the most important metrics for this task.

We consider that there is a trade-off between precision and time response, hence the selection of the best suited models should be made taking the particular application into account. In the cyclists detection, if we look for higher precision and sacrifice some time response, we can choose the Faster R-CNN meta-architecture using InceptionV2 feature extractor. On the other hand, if we are mostly interested in the time response, cyclists detection can be achieved with up to $34$ FPS using SSD-InceptionV2 on a medium-cost computer equipped with a GPU. Similarly, for cyclists orientation detection, our selection is SSD meta-architecture with InceptionV2 feature extractor, since it obtained a good AP for a threshold of $0.5$ over IoU and fast time response, hence it is well suited for the cyclist's orientation detection in real-time embedded applications. However, if we consider precision as a more important feature than speed, we may select Faster R-CNN using InceptionV2 as it obtained similar results in precision than Faster R-CNN with InceptionResNetV2, while proving to be faster than them in response times.
 
In conclusion, the proposed strategy for multi-class orientation detection is a simple but effective alternative way to protect cyclists on the road, by further providing important information about the heading of the cyclist. Furthermore, such information can be easily used along with some tracking algorithm to estimate the cyclist movement, and  predict its trajectory in order to detect potential collisions. The later is left as future task.

Future efforts will be dedicated to extending and balancing the provided dataset, increasing the number of instances by class, but also by size in order to improve the detectors performance, specially for similar classes such as CyclistSE vs CyclistSW. Also,  increasing the number of smaller instances will help  to better detect further away objects.

Finally, it would be interesting to implement and test the proposed techniques embedded on a scaled intelligent vehicle. Also, it is desired to determine the cyclist movement-intention by detecting standard cycling hand signals.

\section*{Acknowledgment}
We thank the Mexican National Council of Science and Technology CONACyT for the grants given. Also to Luis Angel Pinedo Sánchez for contributing to the collection of the images, and ”Sportbike Jerez MTB” for allowing us to take pictures at their events.

\bibliographystyle{iet}
\bibliography{bibilography}

\end{document}